\documentclass[lettersize,journal]{IEEEtran}
\usepackage{amsmath,amsfonts}
\usepackage{algorithmic}
\usepackage{mathrsfs}
\usepackage{algorithm}
\usepackage{array}
\usepackage{textcomp}
\usepackage{stfloats}
\usepackage{url}
\usepackage{verbatim}
\usepackage{graphicx}
\usepackage{cite}
\usepackage{pifont}
\usepackage{url}
\usepackage{color}
\usepackage{tabularx}
\usepackage{multirow}
\usepackage{enumitem}
\usepackage{subcaption}
\usepackage{booktabs}
\usepackage{extarrows}
\usepackage{listings}

\usepackage{amssymb}

\begin{document}

\title{LLM-Assisted Op-Amp Behavioral-Level Design via Agentic Human-Mimicking Reasoning}

\author{Zihao Chen, Ziyi Sun, Jiayin Wang, Ji Zhuang, Jinyi Shen, Xiaoyue Ke,
\\ Li Shang~\IEEEmembership{Member,~IEEE}, Xuan Zeng~\IEEEmembership{Senior Member,~IEEE}, and Fan Yang~\IEEEmembership{Member,~IEEE}

\thanks{This work has been submitted to the IEEE for possible publication. Copyright may be transferred without notice, after which this version may no longer be accessible.}
%\thanks{$^{\dag}$ These authors contributed equally to this paper.}
\thanks{%This work was supported by the National Natural Science Foundation of China (NSFC) Research Project.
The authors are with the State Key Laboratory of Integrated Chips and Systems, Fudan University, Shanghai, China. (Corresponding author: Fan Yang; e-mail: yangfan@fudan.edu.cn.)}
}

\maketitle

\begin{abstract}
This paper proposes \textbf{\emph{White-Op}}, an operational amplifier (op-amp) behavioral-level parameter design framework assisted by the human-mimicking reasoning of large language model agents. 
A symbolic reasoning-numerical solving decoupled paradigm is adopted: the agent performs step-by-step symbolic reasoning and formulates the design as a white-box optimization problem, which is then solved programmatically, verified via simulation, and refined iteratively.
To guide this symbolic design process, implicit human reasoning mechanisms are formalized into explicit steps of introducing hypothetical constraints
during transfer function simplification, pole-zero extraction and position regulation, converting design heuristics into mathematical formulations. 
A programming mapping protocol then standardizes the translation from symbolic designs to executable programs. 
Finally, a causality-driven refinement loop enables the agent to trace simulation–theory mismatches back to specific symbolic reasoning steps and make targeted corrections iteratively until convergence.
Experiments on 9 op-amp topologies demonstrate that White-Op achieves interpretable behavioral-level designs with an average of 8.52\% theoretical prediction error and retains circuit functionality after transistor-level mapping for all topologies, whereas black-box baselines fail in 5 to 7 topologies.
White-Op is open-sourced at \textcolor{blue}{\url{https://github.com/zhchenfdu/whiteop}}.
\end{abstract}
\vspace{-0.5em}
\vspace{-1em}
\section{Introduction} \label{section:intro}
Operational amplifiers (op-amps) are fundamental building blocks in analog circuits. 
Their parameter design directly determines the circuit performance. 
Given the diverse application scenarios and stringent specifications (specs), 
op-amps often require labor-intensive manual parameter customization. 

For op-amps, interpretable behavioral-level (BL) design is crucial. 
BL modeling is concise, enabling an analytical manual design process grounded in mathematical and physical reasoning. 
Such interpretability is not merely a convenience for human understanding, but an analytical basis for maintaining design reliability in downstream implementations against potential deviations.
For example, when the BL design is mapped to the transistor level (TL), 
more complex parasitic effects often introduce performance degradation or even functional failure;
only when the BL design process is sufficiently interpretable,
can designers proactively avoid potential design flaws early on. 
Even when design failure occurs, 
a well-reasoned BL design allows designers to trace the cause and make targeted corrections, 
rather than facing a black-box result that offers no clue for fixing.

Therefore, design assistant tools that offer designers complete, interpretable (\textbf{\emph{white-box}}) reference rationale and processes in a human-like manner
remain highly desirable. %for op-amp BL design.

Traditional analog circuit sizing methods adopt \textbf{\emph{black-box}} methods \cite{GAsizing1, GAsizing2, GAsizing3,BOsizing1, BOsizing2, BOsizing3,RLsizing1, RLsizing2, RLsizing3}, which can find high-score BL parameters, but the pure score-driven search lacks interpretability. 
Furthermore, the over-optimized results pose risks of unforeseen failures in downstream implementation and leave designers with little basis for understanding or refinement.

Recently, large language models (LLMs),
capable of learning knowledge and performing tasks in natural language,
have become promising assistants for analog circuit design \cite{he2025large}. 
Cutting-edge efforts \cite{adollm,llmuso,ledro,easysize,eesizer,liu2025llm,anaflow} mainly integrate LLMs into conventional sizing loops,
where LLMs iteratively adjust parameters based on frequently received simulation feedback and improve circuit performance.
These works show the potential of LLM-assisted analog circuit design, 
but they focus on accumulating design experience rather than generating complete, human-like design rationale,
as pursued in this work.

Achieving our goal first faces a paradigm-level challenge:
\textbf{\emph{mismatch between LLM capabilities and numerical sizing}}. 
Advanced LLMs excel at logic and symbolic reasoning \cite{deepseek-r1, gemini2.5}, 
but are less reliable in precise numerical inference \cite{imani2023mathprompter, dziri2023faith}. 
Since op-amp parameter design is numerical, 
using LLMs directly for parameter tuning does not fully leverage their strengths
and struggles to produce explicit, complete, and traceable design rationale that designers truly need.

A novel LLM-assisted design paradigm is therefore adopted:
\textbf{\emph{decoupling of symbolic reasoning and numerical solving}}. 
Despite the limitation in numerical inference,
LLMs excel at symbolic mathematical reasoning \cite{imani2023mathprompter,dziri2023faith,ahn2024large} and code generation~\cite{zan2023large}. 
This motivates us to decouple the design process into two complementary phases: symbolic design and numerical solving.
The agent conducts step-by-step reasoning to propose a symbolic design, 
i.e., a closed-form, white-box, differentiable optimization problem with constraints. 
The agent then translates this design into automatically executed code.
This paradigm recasts the agent from a numerical tuner into a design reasoner and programmer, fully leveraging its strengths.

However, implementing this symbolic-numerical decoupled design paradigm still faces the following technical challenges:

\emph{\textbf{Implicitness of human symbolic reasoning mechanisms.}}
Beyond unbiased derivation, 
human-like op-amp design heavily relies on implicit symbolic reasoning mechanisms (e.g.,
which approximations are valid,
or which circuit effects are dominant)
that experts routinely apply based on experience and intuition but rarely articulate. 
This implicitness makes it hard to formulate LLM-executable symbolic tasks.

\emph{\textbf{Non-standardization in mapping designs into programs.}}
After formalizing symbolic designs as constrained optimization problems, they must be mapped to programs for numerical solving. 
However, if the agent freely handles implementation details
(e.g., numerical scaling, 
variable abstraction, 
solver configuration, 
output formatting),
inevitable fluctuations across different code-generation attempts
make the numerical solving process unstable and require excessive debugging iterations.

\emph{\textbf{Untraceability of experience-driven trial-and-error loops.}}
Whether driven by black-box sizers or LLMs, 
conventional sizing loops mainly rely on extensive trial-and-error iterations guided by simulation performance. 
While such iterations can get high-score solutions, 
the design rationale usually remains empirical and opaque,
because directly predicting new parameters provides little traceability for diagnosing root causes or constructing the complete design rationale that designers need.

To address these challenges, 
this paper proposes \textbf{\emph{White-Op}}, an op-amp BL design framework assisted by the human-mimicking reasoning of LLM agents. 
Concretely, White-Op operates in a \textbf{\emph{reasoning-solving decoupled design paradigm}}: 
Given a topology and specs, 
the agent first performs step-by-step symbolic reasoning to propose the design as a mathematically constrained optimization problem. 
This formulation is then mapped into executable code via a standardized programming mapping protocol and solved automatically. 
Simulation-theory discrepancies are traced back to 
targeted reasoning steps, and corrections are made iteratively until convergence.

Detailed technical contributions are listed as follows:

\begin{itemize}[leftmargin = 10pt]
\item \textbf{Formalization of symbolic reasoning mechanisms.}
Besides straightforward, unbiased derivation tasks like listing circuit equations and 
solving for the transfer function (TF),
implicit human reasoning mechanisms for harder tasks, 
including TF simplification, pole-zero (PZ) extraction, and PZ positioning,
are distilled into explicit agentic reasoning steps
of \textbf{\emph{introducing hypothetical constraints}} (e.g., ...$\gg$...). 
Thus, these previously intuition-driven tasks are augmented with traceable, verifiable mathematical conditions.

\item \textbf{Standardized programming mapping protocol.}
The protocol provides predefined programming templates that handle critical engineering principles including
unified solver configurations, 
variable scaling preventing extreme numerical values, 
intermediate variable abstraction decomposing complex expressions, 
and fixed program skeleton and interface.
Consequently, symbolic designs are mapped into programs stably, avoiding open-ended programming.

\item \textbf{Causality-driven design refinement loop.}
If theoretical results deviate from simulation ones or do not meet the specs,
the agent diagnoses design flaws, 
traces them back to relevant symbolic reasoning steps, 
and proposes corrections accordingly.
Re-design is then conducted from the target step.
Consequently, the design loop is closed at the reasoning level, 
typically converging within very few iterations and yielding a complete, interpretable design.

\item \textbf{Open-sourced implementation.}
Building such an agentic framework requires substantial engineering effort.
To make it convenient for the community to reproduce, adopt, and further develop, 
White-Op is open-sourced at \textcolor{blue}{\url{https://github.com/zhchenfdu/whiteop}}.
\end{itemize}

Experiments across 9 op-amp topologies (10 trials each) demonstrate that, 
White-Op achieves interpretable BL designs with only 8.52\% theoretical prediction error, 
and retains circuit functionality after TL mapping for all topologies,
unlike black-box baselines that fail in 5 to 7 topologies after mapping.

The rest of this paper is organized as follows:
Section~\ref{s1} introduces related works.
Section~\ref{s2} provides preliminaries.
Section~\ref{s3} introduces White-Op.
Section~\ref{s4} presents experiments and results.
Section~\ref{s5} concludes this paper.
\section{Related Works}\label{s1}

\subsection{Black-Box Sizing Methods}
Black-box sizing methods treat the circuit simulator as an evaluator and search for parameters that optimize performance toward target specs. 
Representative approaches include 
genetic algorithms (GA)~\cite{GAsizing1, GAsizing2, GAsizing3}, 
Bayesian optimization (BO)~\cite{BOsizing1, BOsizing2, BOsizing3}, 
and reinforcement learning (RL)~\cite{RLsizing1, RLsizing2, RLsizing3}. 
These methods differ in search strategies \cite{golzan2025analog}:
population-based heuristics for GA, 
surrogate model-guided sampling for BO, 
and policy network-based trial-and-error for RL, respectively. 
However, they share a common characteristic: 
frequently calling the simulator and tuning parameters based on score feedback
without reasoning steps. 
Consequently, the designs offer little rationale for designers to analyze and refine. 
Over-optimized parameters also risk unforeseen downstream failures.

\subsection{Rule-based Symbolic Derivation Tools}
A branch of works focuses on developing rule-based symbolic derivation software for BL op-amps. 
GPDD~\cite{gpdd} can derive exact symbolic TFs from circuit topologies, but the resulting expressions are lengthy, rendering subsequent PZ extraction difficult. 
Follow-up works~\cite{toposimp, pzsolve} make progress by encoding TF simplification and PZ extraction heuristics into source code, but they lack flexibility and are hard to interpret. 
Furthermore, the flexible regulation of PZ positions to ensure circuit functionality and performance involves more nuanced, context-dependent trade-offs grounded in manual mathematical and physical reasoning.
Consequently, this step remains largely unsolved by predefining algorithmic rules. 

\subsection{LLM-assisted Sizing Methods}
LLMs have recently been introduced into analog circuit parameter design \cite{he2025large}.
Early efforts \cite{artisan,ampagent,llmacd} simply instruct LLMs to apply existing estimation formulas to calculate parameters step by step,
according to knowledge bases containing ready-made parameter calculation recipes for specific topologies.
While this enables the reproduction of known designs, LLMs function primarily as procedural executors and cannot generalize beyond their pre-compiled recipes. 
Recent progress integrates LLMs into simulation-in-the-loop sizing frameworks~\cite{adollm,llmuso,ledro,easysize,eesizer,liu2025llm,anaflow}, where LLMs adjust parameters based on simulation feedback. 
Specifically, references \cite{adollm,llmuso,ledro,easysize} enhance existing black-box sizing frameworks with LLM guidance, while references \cite{eesizer,liu2025llm,anaflow} build sizing workflows directly driven by LLM agents.
Through such loops, LLMs can progressively accumulate experience and improve circuit performance, 
but the underlying design rationale remains opaque and incomplete.
\begin{figure*}[!th]
    \centering
    \begin{subfigure}{.27\textwidth}
    \centering
    \includegraphics[width=\textwidth]{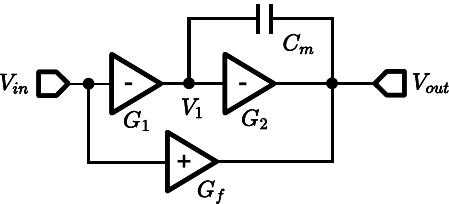}
    \caption{}
    \label{figure:basic_opa_bl}
    \end{subfigure}
    \hspace{4mm}
    \begin{subfigure}{.53\textwidth}
    \centering
    \includegraphics[width=\textwidth]{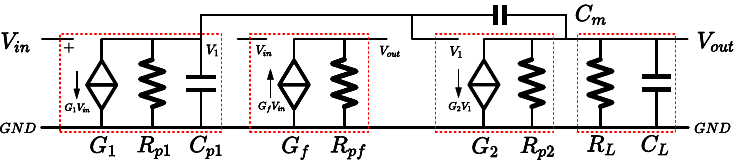}
    \caption{}
    \label{figure:basic_opa_ssm}
    \end{subfigure}
    \vspace{-0.5em}
    \caption{Behavioral-level modeling.
    \textmd{
    Fig.~\ref{figure:basic_opa_bl} is an MZC topology example (loads omitted). 
    Fig.~\ref{figure:basic_opa_ssm} shows the small-signal model.
    }}
\end{figure*}
    
\section{Preliminaries}\label{s2}

\subsection{Op-amp Behavioral-Level Modeling}\label{sec:behavioral-level-modeling}
This work directly handles the BL op-amp.
Fig.~\ref{figure:basic_opa_bl} shows an op-amp example with MZC (multipath zero cancellation) topology \cite{mzc}.
Each stage is
a transconductance $G_i$ with a pair of parasitic resistance $R_{p,i}$ and capacitance $C_{p,i}$ in Fig.~\ref{figure:basic_opa_ssm}.

The transconductance $G_i$ serves as an independent design variable, 
whose sign (- or +) indicates the phase shift (inverting or non-inverting). 

$R_{p,i}$ can be derived from the stage gain $A_i$ and $G_i$:
\begin{equation}
R_{p,i} = {A_i}/{G_i}.\label{2}
\end{equation}

$C_{p,i}$ is actually the input parasitic capacitance of the subsequent stage (null for the final stages connected to the load):
\begin{equation}
C_{p,i} = {G_{i+1}}/{\omega_t},\label{3}
\end{equation}
where $\omega_t \approx 2\pi \cdot f_T$ is the empirical transition frequency value of transistors under the adopted process. 

Besides the transconductance and its gain, 
other design variables include independent resistors and capacitors. 
Loads $R_L$ and $C_L$ are fixed, i.e., not included in the design variables.

\subsection{Problem Formulation} \label{sec:problem-formulation}
For most black-box methods, the op-amp BL parameter design problem is formulated as tuning independent design variables $\mathbf{x}\in\mathcal{X}$ to maximize an objective function $f(\cdot)$, subject to performance specs. 
Formally,
\begin{equation}\label{eq:problem-formulation}
\begin{aligned}
    & \underset{\mathbf{x}\in\mathcal{X}}{\text{max}} 
    & & f(\mathbf{x}) \\
    & \text{\emph{s.t.}} 
    & & \mathbf{y}_\text{min} \le \mathbf{y}(\mathbf{x}) \le \mathbf{y}_\text{max}, \\
\end{aligned}
\end{equation}
where $\mathbf{y}(\mathbf{x})$ is the vector of performance metrics, and $\mathbf{y}_\text{min}$, $\mathbf{y}_\text{max}$ denote the corresponding lower and upper spec bounds. 
For metrics with only a single-sided requirement (e.g., minimum gain or maximum power), the opposite bound can be set to $-\infty$ or $+\infty$ as appropriate. 
The feasible set $\mathcal{X}$ is basically confined by empirical bounds $\mathbf{x}_\text{min}$ and $\mathbf{x}_\text{max}$.

The objective function is set as the figure of merit (FoM):
\begin{equation}
    \text{FoM} = \frac{\text{GBW}[\text{MHz}] \times C_L[\text{pF}]}{\text{Power}[\text{mW}]}.\label{eq:fom}
\end{equation}

Metrics including gain, gain-bandwidth product (GBW), phase margin (PM), and power are considered in the work. 
Note that power is estimated in BL via Eq.~\eqref{eq:power}:
\begin{equation}
\text{Power} = \sum_{i} \frac{G_i}{(g_m/I_D)} \cdot V_{DD}.\label{eq:power}
\end{equation}
where $(g_m/I_D)_i$ is the empirical transconductance-current ratio of the $i$-th amplifier stage,
and $V_{DD}$ is the supply voltage.
% Define example box environment for MZC case study
\newsavebox{\exampleboxcontent}% 定义保存盒子内容的命令
\newenvironment{examplebox}[1][MZC Example]% 定义新环境，可选参数为标题（默认为"MZC Example"）
{%
\begin{lrbox}{\exampleboxcontent}% 开始将内容保存到盒子中
\begin{minipage}{\dimexpr\columnwidth-2\fboxrule-2\fboxsep}% 开始minipage，宽度为列宽减去边框宽度和内边距（避免超出列宽）
\scriptsize% 设置字体大小为scriptsize（比正文小）
\setlength{\abovedisplayskip}{2pt}% 设置显示公式（如equation环境）上方的垂直间距
\setlength{\belowdisplayskip}{2pt}% 设置显示公式下方的垂直间距
\setlength{\abovedisplayshortskip}{0pt}% 设置短显示公式（公式前没有文本）上方的垂直间距
\setlength{\belowdisplayshortskip}{0pt}% 设置短显示公式下方的垂直间距
\setlength{\jot}{1pt}% 设置多行公式（如align环境）行之间的额外间距
\setlength{\leftmargini}{0pt}% 设置第一级列表（itemize/enumerate）的左缩进
\setlength{\leftmarginii}{0pt}% 设置第二级列表的左缩进
\setlength{\parindent}{0pt}% 设置段落首行缩进（0表示不缩进）
\setlength{\itemindent}{0pt}% 设置列表项的缩进
\setlength{\listparindent}{0pt}% 设置列表项内段落的缩进
\setlength{\mathsurround}{0pt}% 设置数学模式（$...$）周围的额外间距
\setlength{\medmuskip}{0mu}% 设置中等数学间距（如关系符号周围的间距，负值使更紧凑）
\setlength{\thickmuskip}{0mu}% 设置粗数学间距（如等号周围的间距，负值使更紧凑）
\setlength{\thinmuskip}{0mu}% 设置细数学间距（如运算符周围的间距，负值使更紧凑）
\setlength{\scriptspace}{0pt}% 设置上下标后的水平间距（使上下标更靠近基字符）
\everymath{% 对每个数学模式应用以下设置
  %\scriptspace=0pt% 设置上下标后的水平间距
  %\fontdimen13\textfont2=2.5pt% 设置上标的垂直位置（使上标更靠近基字符）
  %\fontdimen14\textfont2=2.5pt% 设置下标的垂直位置（使下标更靠近基字符）
}%
\textbf{#1:}\vspace{1pt}\par% 显示标题（加粗）并添加小间距
}%
{%
\end{minipage}% 结束minipage
\end{lrbox}% 结束保存盒子
\par\vspace{4pt}% 在盒子前添加垂直间距
\noindent\fbox{\usebox{\exampleboxcontent}}% 使用fbox绘制边框并显示盒子内容（\noindent避免缩进）
\par\vspace{4pt}% 在盒子后添加垂直间距
}

\section{Proposed Approaches}\label{s3}
This section proposes White-Op,
whose overall paradigm is presented in Section~\ref{s3.1}.
Section~\ref{s3.2} presents 
the symbolic design stage. 
Section~\ref{s3.3} presents 
the numerical solving stage. 
Section~\ref{s3.4} presents
the design refinement stage.

\begin{figure}[!b]
  \centerline{\includegraphics[width=.46\textwidth]{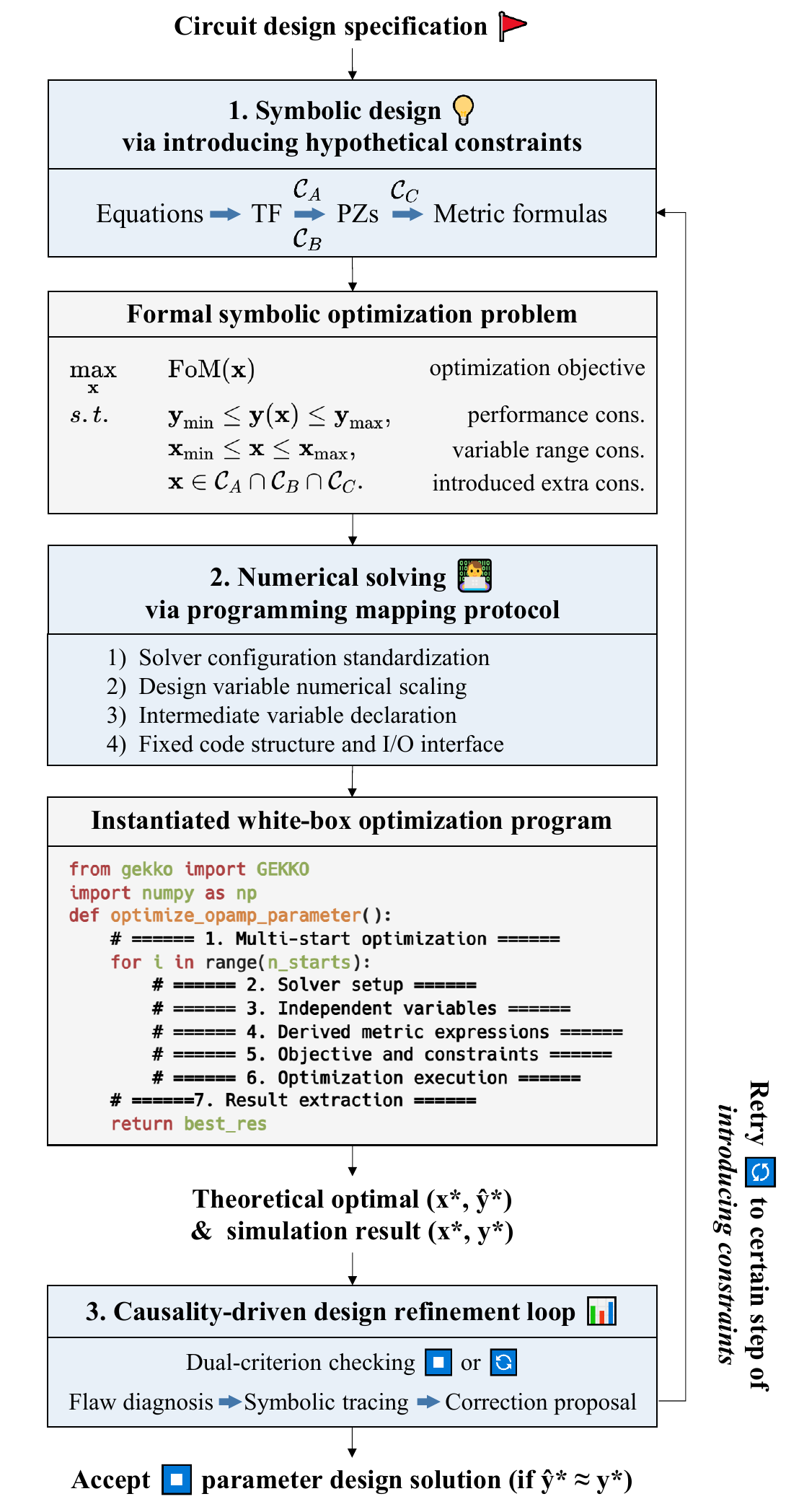}}
  \caption{The overall paradigm of White-Op.
  Note that symbolic reasoning and numerical solving are decoupled in White-Op.}
  \label{fig:workflow}
  \end{figure}

\subsection{Overview of the Symbolic-Numerical Decoupled White-Op}\label{s3.1}
Fig.~\ref{fig:workflow} presents the overall paradigm of White-Op, where symbolic reasoning and numerical solving are decoupled. 
There are three stages in White-Op:
the symbolic design stage centered on introducing hypothetical constraints,
the numerical solving stage based on a programming mapping protocol,
and the causality-driven refinement loop via analyzing design flaws. 
The core ideas of each stage are briefly presented below. 

\subsubsection{\textbf{Symbolic design stage centered on introducing hypothetical constraints}}
The agent leverages its symbolic reasoning capability to derive the symbolic op-amp design. 

The agent lists circuit equations based on the given topology, then derives the unbiased TF with variable substitutions. 

Pivotal challenges arise after obtaining the TF, 
i.e., how to simplify the TF, 
solve for PZ formulas appropriately, and arrange their positions to achieve a valid design. 
Implicit human reasoning mechanisms are distilled into explicit reasoning steps of introducing hypothetical constraints (cons.):

$\bullet$ The unbiased TF is usually too complex to analyze directly; 
thus, it is simplified through coefficient approximation according to extra inequalities (\textbf{\emph{cons. A}}). 

$\bullet$ The simplified numerator and denominator polynomials are then solved. Since direct factorization or root formulas are usually infeasible, extra inequalities (\textbf{\emph{cons. B}}) are introduced to yield concise, analytically meaningful PZ formulas.

$\bullet$ With PZs extracted, the agent introduces inequalities (\textbf{\emph{cons. C}}) to govern the relative positions of PZs, ensuring functional operation. 
Finally, the PZ expressions are used to derive approximate metric formulas. 

This stage outputs a formal symbolic optimization problem and strict reasoning steps to be verified. 
As shown in Fig.~\ref{fig:workflow}, the objective is FoM, 
while the cons. include:
performance cons.,
variable range cons.,
and extra cons. A, B, and C. 

\subsubsection{\textbf{Numerical solving stage based on standardized programming mapping protocol}} 
Given the symbolic optimization problem, 
the agent needs to map it into code for numerical solving based on a standardized programming mapping protocol,
which embeds four essential engineering principles:

$\bullet$ Solver configurations should be unified to ensure the result consistency and the reproducibility of the solving process.

$\bullet$ Variable numerical values should be scaled to certain range to avoid solver convergence issues from extreme values.

$\bullet$ Intermediate variable abstraction should be introduced to decompose overly complex expressions, thereby enhancing code maintainability and reducing coding errors.

$\bullet$ Program structure and interface should be standardized to facilitate result extraction and reduce coding errors.

This protocol establishes a stable programming pipeline. 
As shown in Fig.~\ref{fig:workflow}, 
this stage outputs standardized programs and yields the theoretically predicted optimal $(\mathbf{x}^*, \hat{\mathbf{y}}^*)$ and corresponding simulation results $\mathbf{y}^*$ under $\mathbf{x}^*$.

\subsubsection{\textbf{Causality-driven design refinement loop}} 
After obtaining the theoretical and simulation results, 
the agent checks two aspects: 
whether the metrics meet the specs, 
and whether there is a non-negligible mismatch between the theoretically predicted $\hat{\mathbf{y}}^*$ and the actual $\mathbf{y}^*$.

The failure of the above checks generally means that
the symbolic design is flawed. 
The agent is guided to reason and propose corrective actions like a human designer:

$\bullet$ It first reads the results and diagnoses the observed numerical flaws (e.g., the simulated GBW collapses prematurely). 

$\bullet$ It then traces the flaws back to the original symbolic design process to analyze the violated hypothetical constraint (e.g., the dominant-pole approximation no longer holds).

$\bullet$ It formulates a new corrective proposal (e.g., an alternative hypothetical constraint) and re-designs from the targeted step.

This causality-driven refinement loop typically converges within very few iterations and outputs a complete, interpretable, and analytically sound design trajectory.

To help readers follow our method more concretely, 
the subsequent Sections~\ref{s3.2}, \ref{s3.3}, and \ref{s3.4} 
will elaborate on each stage, 
using the BL parameter design process of the MZC topology in Fig.~\ref{figure:basic_opa_bl} as a running example, 
with key agent responses presented to illustrate the reasoning at each step.

\begin{figure}[!b]
\centerline{\includegraphics[width=.45\textwidth]{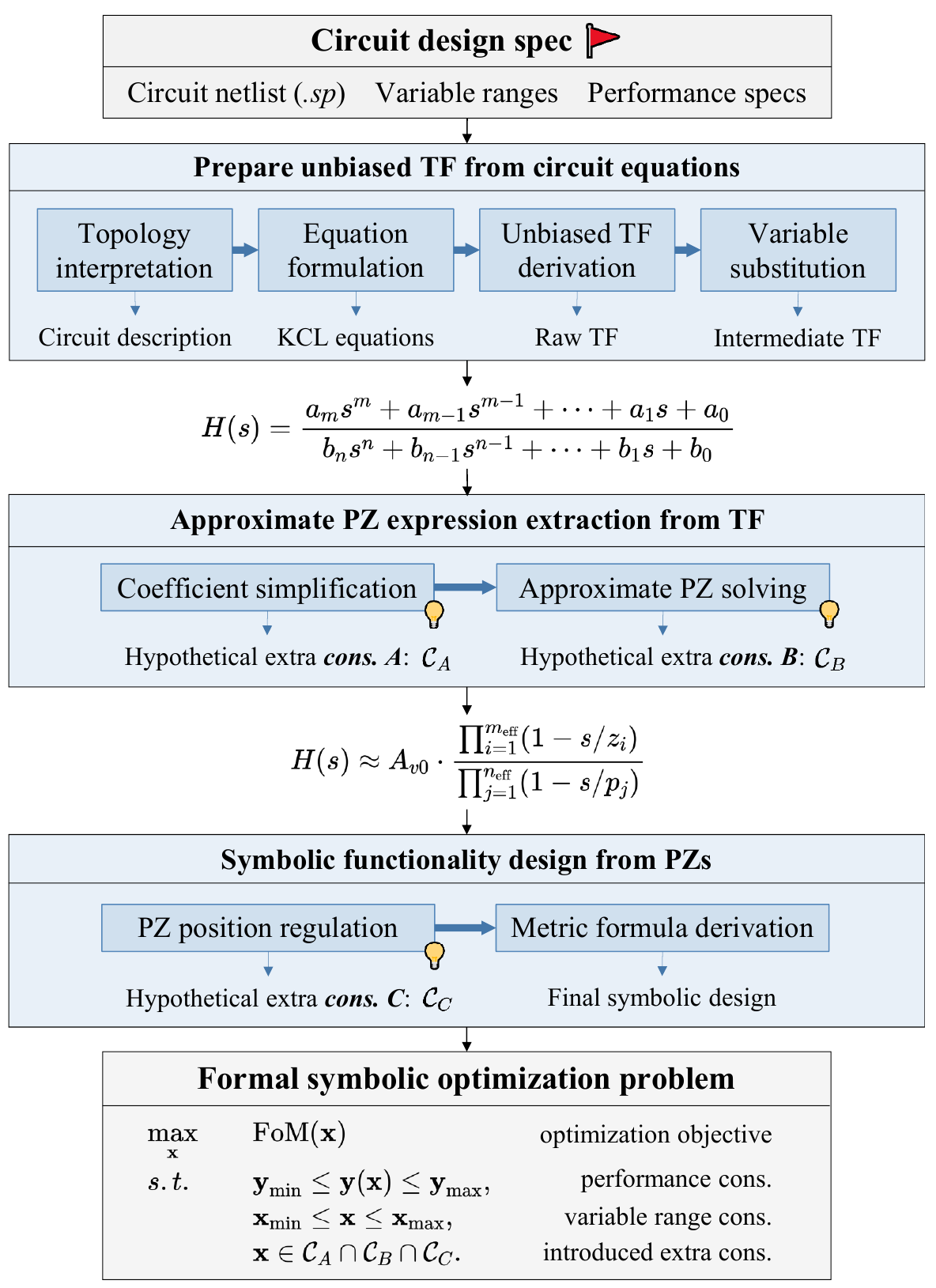}}
\caption{The symbolic design stage centered on introducing hypothetical cons. A, B, C.}
\label{fig:stage1}
\end{figure}

\subsection{Symbolic Design via Introducing Hypothetical Constraints}\label{s3.2}
The agent first performs the symbolic design centered on introducing hypothetical cons., to explicitly embed design thoughts into mathematical conditions. 
As shown in Fig.~\ref{fig:stage1}, the inputs include a netlist, variable bounds, and performance specs. 
This stage comprises three sub-stages: 
deriving the unbiased TF from circuit equations,
extracting PZs from the TF, and 
designing the functionality symbolically from PZs. 

\subsubsection{\textbf{Deriving the unbiased TF from circuit equations}}
The agent begins by listing circuit equations according to the topology and derives the TF symbolically.
The TF is generally a rational expression with both the numerator and denominator being polynomials with real coefficients, expressed as:
\begin{equation} \label{eq:general_tf}
    H(s)  = \frac{a_m s^m + a_{m-1} s^{m-1} + \dots + a_1 s + a_0}{b_n s^n + b_{n-1} s^{n-1} + \dots + b_1 s + b_0},
\end{equation}
where $m$ and $n$ denote the orders of the numerator and denominator polynomials, respectively, and the coefficients $a_i, b_j$ are combinations of the aforementioned design variables.

Here are the detailed substeps:

\textbf{\emph{$\bullet$ Topology interpretation:}} 
The agent is guided to provide a detailed natural language description of all signal paths from input to output. 
It is also required to pay attention to the phase signs of each amplifier stage, 
thereby mitigating careless sign errors in circuit equation formulation.

Below is the example response of the agent's analysis of the MZC op-amp. 
The agent precisely articulates both the small-signal model in Fig. \ref{figure:basic_opa_ssm} and the corresponding signal routing and phase polarities.
For brevity, the original output has been properly abridged and adapted for all following response logs:

\begin{examplebox}[Key response 1, iteration 1 —— Topology description]
Here is the detailed structural interpretation of the circuit.
The circuit is a two-stage op-amp employing Miller compensation, 
enhanced with a feedforward transconductance stage. 
It is single-ended, amplifying the signal from \texttt{NodeInput} to \texttt{NodeOutput}.
  
\textbf{1. First stage (main path)} ...\\
% The signal enters at \texttt{NodeInput}.
% An inverting transconductance stage $G_1$ converts the input voltage into a current flowing from \texttt{Node1} to ground.
% The signal outputs at \texttt{Node1}.
% The total impedance at \texttt{Node1} is determined by:
% the parasitic output resistance of the first stage: $R_{p1}$,
% and the parasitic capacitance $C_{p1}$, which physically represents the input capacitance of the subsequent stage.
\textbf{2. Second stage (main path)} ...\\
% The signal arrives from \texttt{Node1}.
% An inverting transconductance stage $G_2$ amplifies the signal from \texttt{Node1}, driving the current from \texttt{NodeOutput} to ground.
% The signal outputs at \texttt{NodeOutput}.
% Since both stages are inverting, 
% the overall main path provides a non-inverting gain relationship between \texttt{NodeInput} and \texttt{NodeOutput}.
\textbf{3. Feedforward stage} ...\\
% A parallel signal path connects \texttt{NodeInput} to \texttt{NodeOutput} via the transconductance element $G_f$.
% This is a non-inverting transconductance stage,
% which means the feedforward signal is in phase with the signal amplified by the main two-stage path at low frequencies.
% %boosting the overall transconductance or providing a left-half-plane zero (depending on analysis).
% This stage contributes a parasitic resistance $R_{pf}$ to \texttt{NodeOutput}.
\textbf{4. Frequency compensation} ...\\
%A Miller capacitor $C_m$ is connected between \texttt{Node1} (the input of the second stage) and \texttt{NodeOutput} (the output of the second stage)...
%This creates a pole-splitting effect, establishing the dominant pole at \texttt{Node1} and moving the non-dominant pole to a higher frequency
\textbf{5. Output load} ...\\
%The \texttt{NodeOutput} is loaded by a combination of internal parasitics and external components:
%The parallel combination of the second stage parasitic resistance $R_{p2}$, the feedforward stage parasitic resistance $R_{pf}$, and the external load resistor $R_L$.
%The external load capacitor $C_L$.
\textbf{6. Summary of Nodes and Connections}\\
\texttt{NodeInput}: Connected to the inputs of $G_1$ and $G_f$.\\
\texttt{Node1}: Connected to the output of $G_1$, the input of $G_2$, the parasitic impedance of Stage 1 ($R_{p1}, C_{p1}$), and the Miller capacitor $C_m$.\\
\texttt{NodeOutput}: Connected to the outputs of $G_2$ and $G_f$, the Miller capacitor $C_m$, the parasitic resistances ($R_{p2}, R_{pf}$), and the loads.
\end{examplebox}

\textbf{\emph{$\bullet$ Equation formulation:}} 
The agent then formulates the complete set of circuit equations in the $s$-domain using Kirchhoff's Current Law (KCL), to prepare for solving the TF.

In the MZC design example, the agent's key response for this step is shown below,
which closely mirrors the way human designers would manually formulate circuit equations:

\begin{examplebox}[Key response 2, iteration 1 —— KCL equations]
Based on the circuit topology analysis, here are the $s$-domain circuit equations.

\textbf{1. Nodal analysis}

We formulate the KCL equations for \texttt{Node1} and \texttt{NodeOutput}. The sum of currents leaving each node is set to zero.

For \texttt{Node1}, the currents leaving \texttt{Node1} consist of:\\
$\bullet$ The current from the transconductance stage $G_1$: $G_1 V_{in}$,\\
$\bullet$ The current through parasitic resistance $R_{p1}$: ${V_1}/{R_{p1}}$,\\
$\bullet$ The current through parasitic capacitance $C_{p1}$: $s C_{p1} V_1$,\\
$\bullet$ The current through Miller compensation capacitor $C_m$: $s C_m (V_1 - V_{out})$.

For \texttt{NodeOutput}:
The currents leaving \texttt{NodeOutput} consist of:\\
$\bullet$ The current from the feedforward transconductance stage $G_f$: $-G_f V_{in}$,\\
$\bullet$ The current from the second transconductance stage $G_2$: $G_2 V_1$,\\
$\bullet$ The current through parasitic resistance $R_{p2}$: ${V_{out}}/{R_{p2}}$,\\
$\bullet$ The current through feedforward parasitic resistance $R_{pf}$: ${V_{out}}/{R_{pf}}$,\\
$\bullet$ The current through load resistor $R_L$: ${V_{out}}/{R_L}$,\\
$\bullet$ The current through load capacitor $C_L$: $s C_L V_{out}$,\\
$\bullet$ The current through Miller compensation capacitor $C_m$: $s C_m (V_{out} - V_1)$.

\textbf{2. $s$-domain circuit equations}
\begin{equation*}
\begin{cases}
G_1 V_{in} + V_1 (1/R_{p1} + s C_{p1} + s C_m) - V_{out} s C_m = 0 \\
-G_f V_{in} + V_1 (G_2 - s C_m) + V_{out} (1/R_{p2} + 1/R_{pf} + 1/R_L + s C_L + s C_m) = 0
\end{cases}
\end{equation*}
\end{examplebox}

\textbf{\emph{$\bullet$ Unbiased TF derivation:}} 
The agent then solves the KCL equations using Cramer's rule to obtain the standard-form TF, termed the \textit{raw TF},
where every coefficient in the numerator and denominator polynomials is detailed and precise without simplification.
In the MZC design example, the agent performs well in solving the TF as below:
\begin{examplebox}[Key response 3, iteration 1 —— raw TF]
Based on the circuit equations, we solve for the TF $H(s) = V_{out}(s)/V_{in}(s)$.

\textbf{1. Definition of conductance parameters}

To simplify the expression, we define the total output conductance at \texttt{Node1} ($g_{o1}$) and \texttt{NodeOutput} ($g_{o2}$):
$g_{o1} = {1}/{R_{p1}}$ and 
$g_{o2} = {1}/{R_{p2}} + {1}/{R_{pf}} + {1}/{R_L}$.

\textbf{2. System of equations in matrix form}

Rearranging the KCL equations:
\begin{equation*}
\begin{cases}
(g_{o1} + s(C_{p1} + C_m)) V_1 - s C_m V_{out} = -G_1 V_{in} \\  
(G_2 - s C_m) V_1 + (g_{o2} + s(C_L + C_m)) V_{out} = G_f V_{in}
\end{cases}
\end{equation*}

\textbf{3. Solving for the TF}

Using Cramer's Rule, the determinant of the system matrix (Denominator $D(s)$) is ...
% \begin{align*}
% D(s) = ...
% = g_{o1}g_{o2} + s[g_{o1}(C_L + C_m) + g_{o2}(C_{p1} + C_m) + G_2 C_m] + s^2[C_L C_{p1} + C_L C_m + C_m C_{p1}]
% \end{align*}

The numerator determinant for $V_{out}$ (Numerator $N(s)$) is ...
% \begin{align*}
% N_{out}(s) = [g_{o1} + s(C_{p1} + C_m)][G_f V_{in}] - [-G_1 V_{in}][G_2 - s C_m] \\
% N(s) = V_{out}/V_{in} \cdot D(s) = G_f [g_{o1} + s(C_{p1} + C_m)] + G_1 (G_2 - s C_m) \\
% N(s) = (G_1 G_2 + G_f g_{o1}) + s [G_f(C_{p1} + C_m) - G_1 C_m]
% \end{align*}
% $$N(s) = (G_1 G_2 + G_f g_{o1}) + s [G_f(C_{p1} + C_m) - G_1 C_m]$$

\textbf{4. Final TF}\\
Expanding the conductance terms back to their resistor forms:
$$H(s) = {(a_1 s + a_0)}/{(b_2 s^2 + b_1 s + b_0)}$$
where the coefficients are:
\begin{align*}  
a_1 &= G_f (C_{p1} + C_m) - G_1 C_m, \\
a_0 &= G_1 G_2 + {G_f}/{R_{p1}},\\
b_2 &= C_L C_{p1} + C_L C_m + C_m C_{p1}\\
b_1 &= C_m G_2 + \left( {1}/{R_{p2}} + {1}/{R_{pf}} + {1}/{R_L} \right) (C_{p1} + C_m) + {1}/{R_{p1}} (C_L + C_m), \\
b_0 &= {1}/{R_{p1}} \left( {1}/{R_{p2}} + {1}/{R_{pf}} + {1}/{R_L} \right).
\end{align*}
%\vspace{0.3em}
%where $\omega_t = 2\pi \times 200$ MHz.
\end{examplebox}

\textbf{\emph{$\bullet$ Variable substitution:}}
Following Section \ref{sec:behavioral-level-modeling}, transconductances $G$, stage gains $A$, independent resistances $R$ and capacitances $C$ are used as design variables.
Parasitic terms are substituted per Eqs.~\eqref{2} and \eqref{3}, yielding the \textit{intermediate TF}.
In the MZC design example, the agent's key responses for this step are shown below:

\begin{examplebox}[Key response 4, iteration 1 —— Intermediate TF]
...
\begin{equation*}
H(s) = {(a_1 s + a_0)}/{(b_2 s^2 + b_1 s + b_0)}.
\end{equation*}

After substitution ($R_{p1} = A_1/G_1$, $C_{p1} = G_2/\omega_t$, $R_{p2} = A_2/G_2$, $R_{pf} = A_f/G_f$), we get the intermediate TF coefficients as follows:
\begin{align*}
a_1 &= G_f (C_{p1} + C_m) - G_1 C_m = G_f G_2/\omega_t + C_m (G_f - G_1), \\
a_0 &= G_1 G_2 + G_f/R_{p1} = G_1 G_2 + G_1 G_f/A_1, \\
b_2 &= C_L C_{p1} + C_L C_m + C_m C_{p1} = (G_2/\omega_t)(C_L + C_m) + C_L C_m, \\
b_1 &= G_2 C_m + (G_2/A_2 + G_f/A_f + 1/R_L) (G_2/\omega_t + C_m) + (G_1/A_1) (C_L + C_m), \\
b_0 &= (1/R_{p1}) (1/R_{p2} + 1/R_{pf} + 1/R_L) = (G_1/A_1) (G_2/A_2 + G_f/A_f + 1/R_L).
\end{align*}
\end{examplebox}

\subsubsection{\textbf{Approximate PZ extraction from TF}} \label{sec:approximate-pz-extraction}
The TF above is too complex to analyze directly.
The agent usually simplifies the TF first and then extracts approximate yet traceable PZ formulas.
This step specifically exemplifies human-like reasoning with implicit intuitions and experiences such as why some terms can be simplified or even omitted but other terms should be retained, how to determine the threshold of simplification, etc.
Facing such implicit reasoning rationale, 
the agent is guided to introduce \textbf{\emph{extra cons.}} to ensure the validity of simplifications.
This conditional factorization process can be formalized by the following expression, where the polynomial is approximated from Eq.~\eqref{eq:general_tf} into a root-extracted form:
\begin{equation} \label{eq:pz_extraction}
  H(s) = \frac{\sum_{i=0}^{m} a_i s^i}{\sum_{j=0}^{n} b_j s^j} 
  \mathrel{\underset{\text{cons. B}}{\overset{\text{cons. A}}{\scalebox{3.5}[1]{$\approx$}}}}
  A_{v0} \cdot \frac{\prod_{i=1}^{m_\text{eff}} \left(1 - \frac{s}{z_i}\right)}{\prod_{j=1}^{n_\text{eff}} \left(1 - \frac{s}{p_j}\right)},
\end{equation}
where $A_{v0} = a_0/b_0$ represents the gain, 
while $z_i$ and $p_j$ denote the approximated zeros and poles, respectively. 
Note that $m_\text{eff} \le m$ and $n_\text{eff} \le n$, 
because the agent sometimes safely ignores extremely high-frequency roots. 
The operator $\mathrel{\underset{\text{cons. B}}{\overset{\text{cons. A}}{\scalebox{3.5}[1]{$\approx$}}}}$ explicitly dictates that this symbolic factorization strictly holds only when the introduced extra cons. A and B are satisfied.

Specifically, this step is decomposed into two substeps:

\textbf{\emph{$\bullet$ Coefficient simplification:}} \label{sec:coefficient-simplification}
The agent performs simplifications on polynomial coefficients based on variable bounds to facilitate root-finding.
When simplifying, it should balance necessity (whether the term's complexity hinders subsequent derivation) against feasibility (whether the simplification is robustly valid within variable bounds).
Specifically, the agent (i) simplifies the TF through a symbolic dominance analysis based on parameter orders of magnitude; 
(ii) then selectively drops negligible terms while preserving critical characteristics, e.g., left-half-plane (LHP) or right-half-plane (RHP) of PZs.
Any mathematical hypothesis justifying a simplification is noted as an extra constraint, i.e., \textbf{\textit{cons. A}}.

Note that parasitic capacitances could be comparable to or even larger than compensation capacitances; therefore, the agent should carefully consider them (e.g., explicitly using equivalent total node capacitance in derivation).
Besides, inequality cons. are generally preferred over equality cons., as the latter may be fragile during downstream implementation and may significantly reduce the design space.
In the MZC design example, the agent's coefficient simplification strategies are summarized below:

\begin{examplebox}[Key response 5, iteration 1 —— Coefficient simplification]
We define the following auxiliary variables to represent physical conductances and capacitances derived from design variables:
stage 1 output conductance is $g_{o1} = {G_1}/{A_1}$,
stage 2 output conductance is $g_{o2} = {G_2}/{A_2} + {G_f}/{A_f} + {1}/{R_L}$, 
and stage 2 input parasitic capacitance is $C_{p1} = {G_2}/{\omega_t}$.

Now we simplify the coefficients $a_i$ and $b_i$ based on the design variable ranges...

\textbf{1. Numerator:}\\
For $a_0 = G_1 G_2 + {G_1 G_f}/{A_1}$:
we compare $G_1 G_2$ with ${G_1 G_f}/{A_1}$.
Since $A_1 \ge 40$, the second term is $< 2.5\%$ of the first term (assuming $G_f \approx G_2$).
Therefore we have $a_0 \approx G_1 G_2$.
Finally we introduce \textbf{cons. A1}: $A_1 \gg 1$ (naturally satisfied).

For $a_1 = G_f (C_{p1} + C_m) - G_1 C_m$:
no simplification is performed here to preserve the ability to model the RHP zero cancellation or LHP zero creation by $G_f$.

\textbf{2. Denominator:}\\
For $b_0 = g_{o1} g_{o2}$: we keep its exact form.

For $b_1 = G_2 C_m + g_{o2}(C_{p1} + C_m) + g_{o1}(C_L + C_m)$:
we compare the term $G_2 C_m$ with $g_{o2}(C_{p1} + C_m) + g_{o1}(C_L + C_m)$.
Since $g_{o2}$ is small ($\mu \text{S}$), and $C_{p1}, C_m$ are small ($\text{pF}$), so $g_{o2}(C_{p1} + C_m)$ is negligible.
Therefore we have $b_1 \approx G_2 C_m$.
To ensure Miller term dominates output pole interaction, we introduce \textbf{cons. A2}: $C_m > {G_1 C_L}/{G_2 A_1}$. 

For $b_2 = C_L C_{p1} + C_L C_m + C_m C_{p1}$:
we compare $C_L = 10 \text{pF}$ with $C_{p1} = G_2/\omega_t \approx 0.1 ~\text{to}~ 0.8 \text{pF}$.
Thus $C_L \gg C_{p1}$.
Therefore we have $b_2 \approx C_L (C_m + C_{p1})$.
We retain $C_{p1}$ which is crucial when the optimizer minimizes $C_m$.
Finally we introduce \textbf{cons. A3}: $C_L \gg C_{p1}$, which is satisfied by design $C_L=10\text{pF}$ vs parasitic $\text{fF}$.
\end{examplebox}
  
\textbf{\emph{$\bullet$ Approximate PZ solving:}} 
The agent is guided to extract PZs from the simplified polynomials derived previously. 
Since exact algebraic solutions are mathematically intractable and lack design intuition, the agent is encouraged to fully exercise its mathematical capabilities to customize solving strategies. 
The extra cons. in this step are denoted as \textbf{\textit{cons. B}}.
The agent usually follows a condition-driven reasoning process:

\textbf{(i) Direct factorization:} 
If trivial algebraic factorization is achievable (e.g., 1st-order polynomial $N(s) = a_1 s + a_0 = 0$ in the example below), the agent extracts roots directly. However, this ideal scenario is rare.

\textbf{(ii) High-order deflation:} 
For 3rd-order (or higher) polynomials, exact root formulas are impractical. 
The agent assumes one dominant low-frequency root (e.g., $p_{\text{dom}} \approx -b_0/b_1$). 
By factoring out $(1 - s/p_{\text{dom}})$, it successfully reduces the original polynomial into a 2nd-order expression for further analysis.

\textbf{(iii) Quadratic root analysis:} 
For a 2nd-order polynomial (e.g., $b_2 s^2 + b_1 s + b_0 = 0$), the agent should first estimate the discriminant ($b_1^2 - 4 b_0 b_2$) based on typical design values to determine the suitable strategy. 
The agent can introduce $b_1^2 \gg 4 b_0 b_2$ and apply the \textbf{\emph{real root separation}} strategy with widely separated roots, i.e., $p_1 \approx -b_0/b_1$ and $p_2 \approx -b_1/b_2$.
The agent can also set $b_1^2 \ll 4 b_0 b_2$ (which often happens in some complex compensation structures) and apply the \textbf{\emph{complex conjugate root extraction}}, utilizing natural frequency $\omega_n = \sqrt{b_0/b_2}$ and damping ratio $\zeta = b_1 / (2\sqrt{b_0 b_2})$ to express the complex conjugate roots.

The following example demonstrates how the agent executes the real root separation technique for the denominator, and direct factorization for the numerator of the MZC example:

\begin{examplebox}[Key response 6, iteration 1 —— Approximate PZ solving]
\textbf{1. Zeros from $N(s) = 0$:}\\
The zero is located at $z_1 = -a_0 / a_1$.
$$z_1 = - \frac{G_1 G_2}{G_f C_{p1} + C_m (G_f - G_1)}$$
If $G_f = 0$, $z_1 = G_2/C_m$ (RHP). If $G_f$ is large enough such that the denominator becomes positive, the zero moves to the LHP.

\textbf{2. Poles from $D(s) = 0$:}\\
We assume widely separated poles to factor 
$D(s) \approx b_2 (s + {b_0}/{b_1}) (s + {b_1}/{b_2})$,

For the dominant pole ($p_1$):
$$p_1 \approx - \frac{b_0}{b_1} = - \frac{g_{o1} g_{o2}}{G_2 C_m}$$
We introduce \textbf{cons. B1}: $b_1^2 \gg 4 b_0 b_2$. Both roots should be real and widely separated. 
This is the condition for a stable operational amplifier design.

For the non-dominant pole ($p_2$):
  $$p_2 \approx - \frac{b_1}{b_2} = - \frac{G_2 C_m}{C_L (C_m + C_{p1})}$$
Note that as $C_m \to \infty$, $p_2 \to -G_2/C_L$; as $C_m \to 0$, $p_2$ is limited by $C_{p1}$.
\end{examplebox}

\subsubsection{\textbf{Symbolic functionality design from PZs}} \label{sec:design-from-pzs}
PZ positions must be carefully regulated to ensure system stability. 
Mathematically, the exact metric expressions, denoted as $\mathbf{y}_{\text{exact}}(\mathbf{x})$, are hard to derive precisely; even if obtained, their complex forms are usually too difficult to analyze. 
To formalize the symbolic design process, $\mathbf{y}(\mathbf{x})$ denotes the simplified, approximate formulas. The approximation relation is written as:
\begin{equation} \label{eq:metric_reduction_abstract}
\mathbf{y}_{\text{exact}}(\mathbf{x}) 
\mathrel{\underset{\text{\textbf{cons. C}}}{\overset{\text{\textbf{cons. A, B}}}{\scalebox{5}[1]{$\approx$}}}} 
\mathbf{y}(\mathbf{x}) = [\text{Gain}, \text{GBW}, \text{PM}, \text{Power}]^\top.
\end{equation}
The above extra constraints are necessary to ensure that the approximated expressions in $\mathbf{y}(\mathbf{x})$ are valid.
  
\textbf{\emph{$\bullet$ PZ positioning:}} 
As shown in Eq.~\eqref{eq:metric_reduction_abstract}, to ensure reasonable values of metrics like PM, \textbf{\textit{cons. C}} are introduced to regulate the mathematical properties of PZs.

Especially, the agent
(i) should explicitly enforce fundamental analog design principles, such as maintaining a single-pole roll-off characteristic within the GBW; 
(ii) should carefully consider the inequality operators (e.g., $\gg$) for numerical calibration. 
If the margin multiplier is either overly aggressive or excessively relaxed, it may mathematically lead to infeasible space or unexpected PZ entanglement. 
(iii) Consistent with Section \ref{sec:coefficient-simplification}, 
inequality cons. are generally preferred over equality cons. for PZ positioning. 

Interestingly, in the following MZC design example, the agent correctly provides the zero-positioning strategy, but overlooks the single-pole roll-off assumption, demonstrating the importance of the iterative mechanism.

\begin{examplebox}[Key response 7, iteration 1 —— PZ position regulation]
Based on the TF and the circuit topology, 
we analyze the singularities under the design variable ranges...

\textbf{1. System Order:} 
The simplified transfer function is a 2nd-order system with one zero ($z_1$) and two poles ($p_1, p_2$).

\textbf{2. Zeros ($z_1$):}
The position depends critically on the sign of $(G_f - G_1)$.
If $G_f < G_1$, the denominator can be negative, leading to a RHP zero, which degrades phase margin.
If $G_f > G_1$, the denominator is positive, yielding a LHP zero. This LHP zero provides phase lead, which is beneficial for compensation.
Therefore, to prevent RHP zero induced phase lag and simplify stability analysis, 
we force the feedforward path to be stronger than the main path's Miller feedthrough.
\textbf{cons. C1}: $G_f \ge 1.1 \cdot G_1$.

\end{examplebox}

\textbf{\emph{$\bullet$ Metric formula derivation:}} 
The agent can finally derive analytical formulas of performance metrics including Gain, GBW, PM, and Power
based on the cooked TF and regulated pole-zero parameters. 
The cooked formulas in this MZC design example (iteration 1) are shown as follows:

\begin{examplebox}[Key response 8, iteration 1 —— Metric formula derivation]
Using auxiliary variables $g_{o1} = G_1/A_1$, $g_{o2} = G_2/A_2 + G_f/A_f + 1/R_L$, and $C_{p1} = G_2/\omega_t$, we obtain the following formulas:

PZs (rad/s):
$$z_1 = \frac{G_1 G_2}{G_f C_{p1} + C_m (G_f - G_1)}, \quad p_2 = \frac{G_2 C_m}{C_L (C_m + C_{p1})}, \quad p_1 = \frac{g_{o1} g_{o2}}{C_m G_2};$$

PM (degrees):
$$\text{PM} = 90^\circ - \frac{180^\circ}{\pi}\arctan\left(\frac{\text{GBW}}{|p_2|}\right) + \frac{180^\circ}{\pi}\arctan\left(\frac{\text{GBW}}{|z_1|}\right);$$

Gain (dB):
$$\qquad\text{Gain}_\text{dB} = 20\cdot\log_{10}(\text{Gain})= 20 \cdot\log_{10}\left(\frac{G_1 G_2}{g_{o1} g_{o2}}\right);$$

GBW (Hz):
$$\qquad\text{GBW} =\text{Gain} \cdot \frac{|p_1|}{2\pi} = \frac{G_1}{2\pi C_m};$$

Power (W):
$$\qquad\text{Power} =  \frac{V_{DD}}{(g_m/I_d)}(G_1 + G_2 + G_f).$$
\end{examplebox}

\subsubsection{Summary}
This stage finally combines all derived formulas and hypothetical conditions into a white-box, differentiable optimization problem.
Different from the problem formulation in Eq.~\eqref{eq:problem-formulation}, 
where $\mathbf{y}(\mathbf{x})$ is no longer evaluated by a black-box simulation, but computed from the derived symbolic expressions under cons. A, B, and C.
Accordingly, the original empirical bounds are further tightened by these constraints. Let $\mathcal{C}_{A}$, $\mathcal{C}_{B}$, and $\mathcal{C}_{C}$ be the feasible regions induced by these cons. respectively,
the white-box optimization problem is written as:
\begin{equation}\label{eq:white_box_problem}
\begin{aligned}
\underset{\mathbf{x}}{\text{max}} \quad
& \mathrm{FoM}(\mathbf{x})\\
{s.t.}\quad
& \mathbf{y}_{\min}\le \mathbf{y}(\mathbf{x})\le \mathbf{y}_{\max},\\
& \mathbf{x}_{\min}\le \mathbf{x}\le \mathbf{x}_{\max},\\
& \mathbf{x}\in\mathcal{C}_{A}\cap\mathcal{C}_{B}\cap\mathcal{C}_{C}.
\end{aligned}
\end{equation}

\subsection{Numerical Solving via Programming Mapping Protocol}\label{s3.3}
The second stage of White-Op turns the symbolic design in 
Eq.~\eqref{eq:white_box_problem} into a numerical optimization program. 
This step is not left to free-form code generation, which frequently produces unexpected bugs and unstable performances. 
To make this mapping stable and reproducible, 
a strict programming mapping protocol is organized around four principles:

\subsubsection{\textbf{Solver configuration unification}}
\texttt{IPOPT} \cite{ipopt} serves as the solver, 
which efficiently solves nonlinear programming problems via a primal-dual interior-point method. 
It is well suited to our scenario where constraints often define narrow and non-convex feasible regions.

To avoid performance variation, hyperparameters are fixed to uniform values. 
Note that as \texttt{IPOPT} is a local optimizer, 
the protocol enforces a multi-start strategy, i.e., 
initial points are sampled from the design space, 
each initial point is solved independently, 
and the optimal solution is retained. 

\subsubsection{\textbf{Numerical scaling for design variables}}
Circuit design variables often span many orders of magnitude. 
If such quantities are written directly in SI units, their products or ratios can become extremely small or large (e.g., $10^{-16}$ in magnitude), 
which makes the optimization problem poorly conditioned with unhealthy Hessian and Jacobian matrices.

Therefore, the protocol mandates all design variables be declared in engineering-scaled units, 
so that their values remain within $[10^{-3},\,10^3]$. 
Typical choices include $\mu$S for transconductance, pF for capacitance, and k$\Omega$ for resistance. 

The same principle also applies to constraints. 
Whenever a cons. contains products or ratios whose SI values would be numerically extreme, 
the cons. should be written in scaled form to stay within a healthy numerical range.

\subsubsection{\textbf{Intermediate variables for complex expressions}}
Many formulas derived in Section~\ref{s3.2} contain nested nonlinear terms, 
such as $\arctan(\cdot)$ in PM or $\log_{10}(\cdot)$ in gain-in-dB. 
In principle, such terms could be written inline. 
In the automated programming pipeline, 
however, deeply nested expressions are easier to mistranslate than decomposed ones, and more difficult to be detected by the agent when a program fails.

Therefore, the protocol requires every derived formula,
especially nonlinear sub-expressions, 
to be introduced as an explicit named intermediate variable. 
It keeps a clear correspondence between the symbolic formulas and the executable code, 
to reduce implementation errors. 

\begin{figure}[!b]
  \centerline{\includegraphics[width=.45\textwidth]{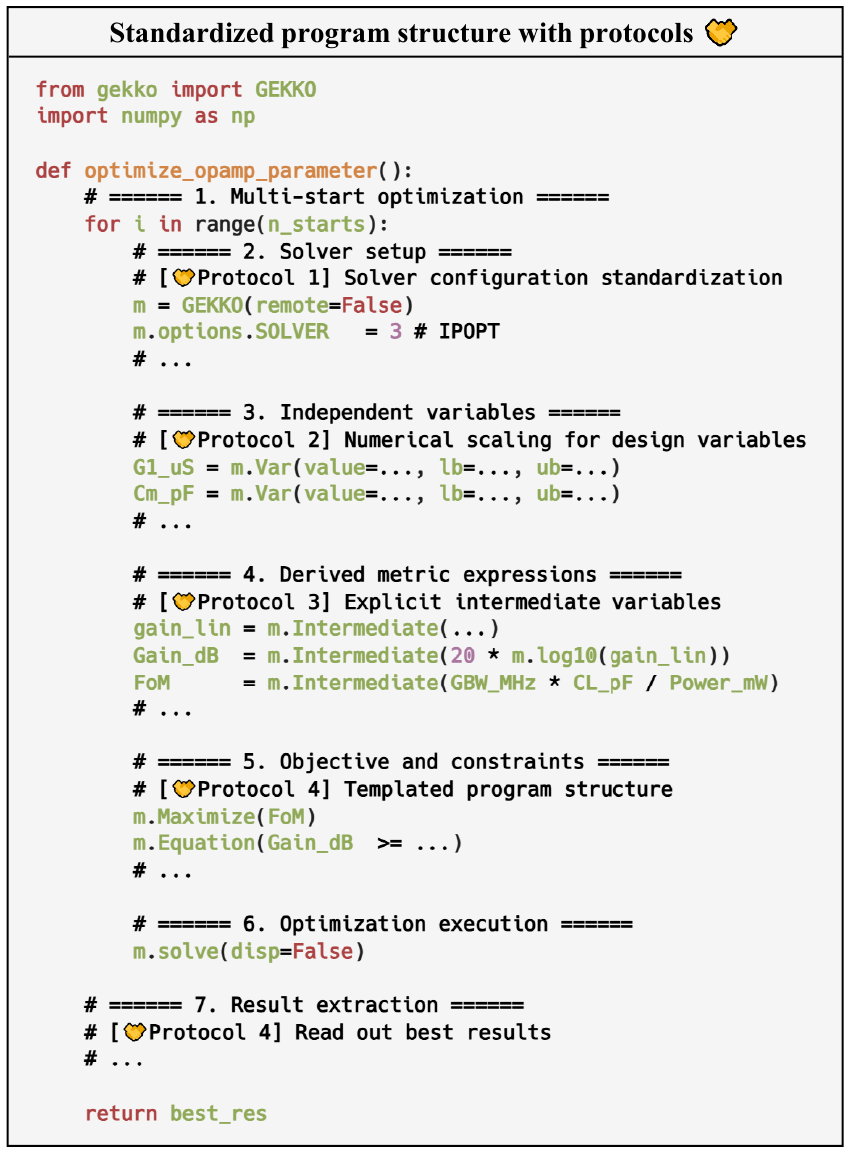}}
  \caption{The standardized program structure in the protocol.}
  \label{fig:protocol_template}
  \end{figure}

\subsubsection{\textbf{Standardized code structure and interface}}
Finally, the protocol fixes the program structure as shown in Fig.~\ref{fig:protocol_template}. 
Every program follows the same sequence: 
multi-start execution, solver setup, scaled variable declaration, derived formulas construction, objective and constraints, optimization execution, and result outputs. 
Only the design-specific contents are filled in by the agent, e.g., variable bounds, algebraic relations, derived formulas in Section~\ref{s3.2}, and any introduced constraints.

\subsubsection{Summary}
As shown in Fig.~\ref{fig:workflow},
this stage delivers two categories of outputs. 
The first is the numerical optimization result: the optimal 
design variable $\mathbf{x}^*$ and its corresponding theoretical
performance metrics $\hat{\mathbf{y}}^*$. 
The second is the simulation result: $\mathbf{x}^*$ is 
substituted into the circuit netlist and evaluated by simulator, 
yielding ground-truth metrics $\mathbf{y}^*$.

\subsection{Causality-driven Design Refinement Loop}\label{s3.4}
\begin{figure}[!b]
\centerline{\includegraphics[width=.45\textwidth]{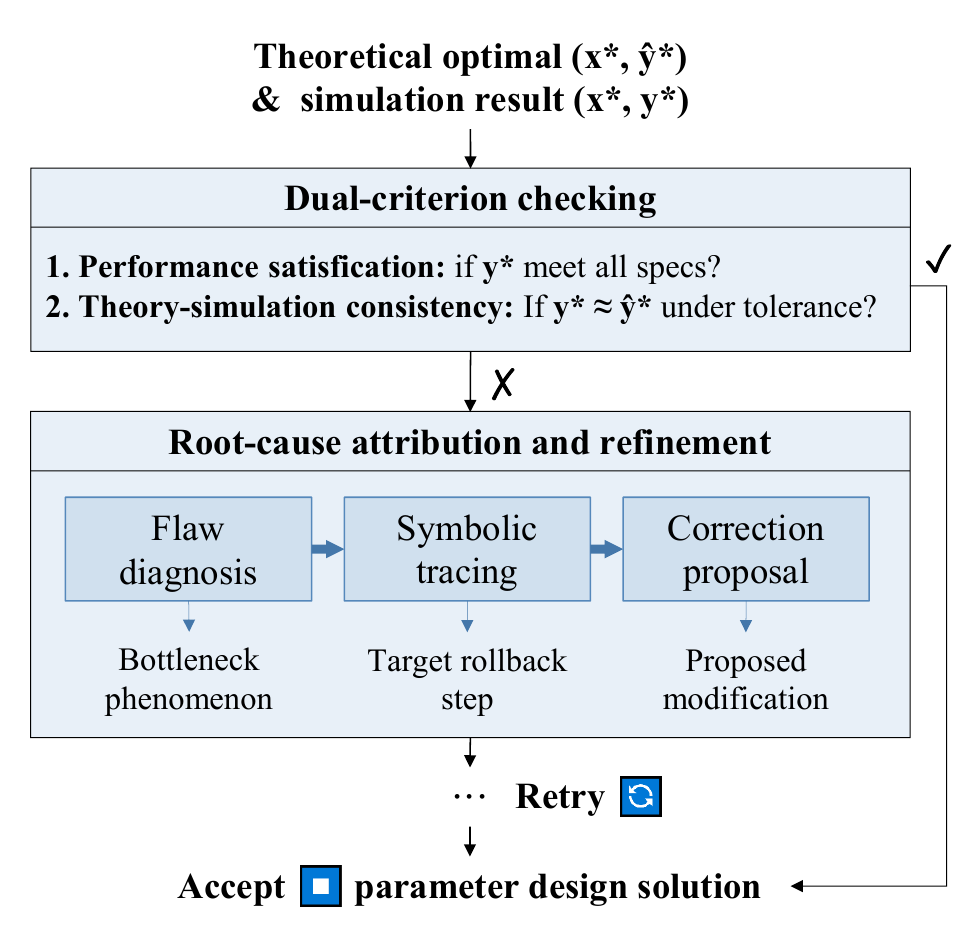}}
\caption{The causality-driven design refinement loop.}
\label{fig:stage3}
\end{figure}

As shown in Fig.~\ref{fig:stage3},
White-Op closes the design loop at the reasoning level, 
enabling a causality-driven refinement loop. 

\subsubsection{\textbf{Dual-criterion checking}}
Upon receiving the simulation results $\mathbf{y}^*$ for the current design point $\mathbf{x}^*$, the agent performs a structured dual-criterion evaluation:

\textbf{\emph{$\bullet$ Performance satisfaction:}} whether all target performance metrics (gain, GBW, PM, power) meet the design specs.

\textbf{\emph{$\bullet$ Theory-simulation consistency:}} whether the 
discrepancy between simulated $\mathbf{y}^*$ and theoretical $\hat{\mathbf{y}}^*$ is within tolerance.

When the discrepancy between $\hat{\mathbf{y}}^*$ and $\mathbf{y}^*$ is non-negligible, 
or some metrics fail to meet the specs, 
this information is interpreted as a diagnostic signal, indicating that the previous symbolic design is flawed and needs to be refined.

\subsubsection{\textbf{Cause attribution and refinement}}
When the above checking fails, the agent executes a causal reasoning process:

\textbf{\emph{$\bullet$ Flaw diagnosis:}} 
The agent first analyzes the current design failure.
For example, when the white-box numerical optimization problem fails to solve, the agent can attribute it to a constraint that is too strict, resulting in an empty feasible region.
When the simulated GBW is significantly lower than the theoretical GBW, the agent can attribute it to an unexpected gain roll-off phenomenon.

\textbf{\emph{$\bullet$ Symbolic tracing:}} 
The agent traces the symptom back through the symbolic derivation chain to identify the violated hypothetical constraints in certain steps. 
For the MZC example (iteration 1), the optimizer, trying to maximize GBW, implicitly violated the condition that $|p_2| > \text{GBW}$, which was an unstated but crucial assumption for the single-pole GBW formula in the PZ positioning step.

\textbf{\emph{$\bullet$ Correction proposal:}} 
The agent formulates a corrective constraint (e.g., an explicit bound on the parasitic pole location) and rolls back to the target step. 
For the MZC example (iteration 1), this leads to introducing cons. C3: $|p_2| \ge 2 \cdot \text{GBW}$ to explicitly enforce the necessary pole separation.

\begin{figure}[!h]
  \centerline{\includegraphics[width=.35\textwidth]{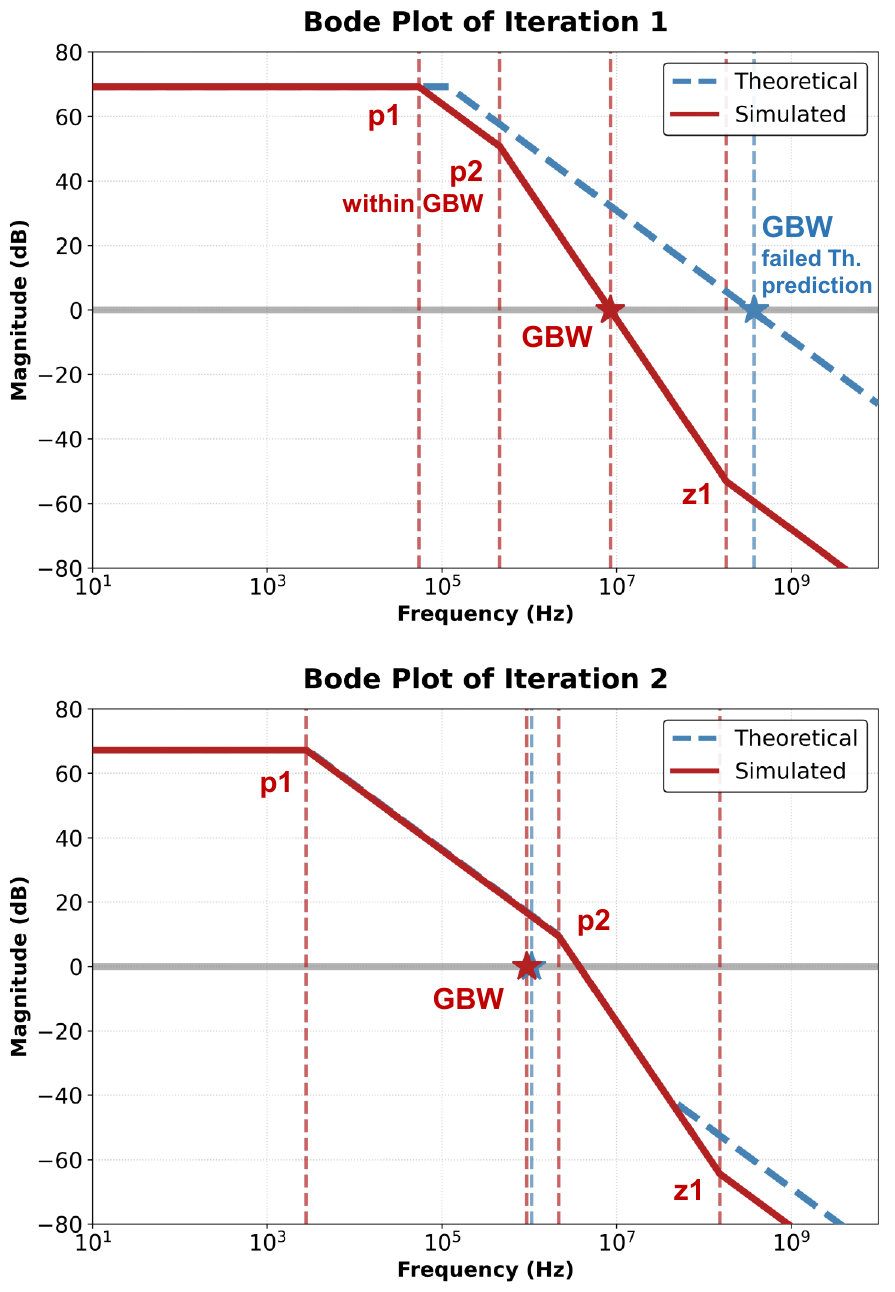}}
  \caption{The Bode plot of the MZC design example.}
  \label{fig:bode_plot}
  \end{figure}

The following details two iterations of the decision-making process in the MZC example.
In iteration 1, the agent's analysis regarding the cause of the failure is correct. 
As shown in Fig.~\ref{fig:bode_plot}, the discrepancy between the theoretical prediction and the simulated results is obvious.
In the initial design, it is implicitly assumed that $p_2$ would be larger than GBW, 
which is not explicitly included as an extra constraint. 
Consequently, the theoretical GBW was calculated using the formula $\text{GBW} = G_1/(2\pi C_m)$, leading to the idealized blue dashed-line Bode plot shown in the first panel of Fig.~\ref{fig:bode_plot}.

However, the actual optimization exploited this implicit assumption. 
By attempting to minimize $C_m$ to increase GBW, a side effect occurred: $p_2$ was pushed inside the GBW. 
This caused the gain roll-off slope to transition rapidly from -20~dB/dec to -40~dB/dec, resulting in a simulated GBW significantly lower than the theoretical value. 
Thus, PM is degraded to only 6.1$^\circ$, rendering the circuit non-functional.

Therefore, the agent proposes to roll back and introduces cons. C3: $|p_2| \ge 2 \cdot \text{GBW}$. 
In iteration 2, the alignment between theory and simulation validates the effectiveness of the correction, as shown in the second panel of Fig.~\ref{fig:bode_plot}.

\begin{examplebox}[Key response 9, iteration 1 —— Error analysis]
  \begin{center}
{%
  \setlength{\tabcolsep}{2pt}
  \renewcommand{\arraystretch}{0.7}
  \begin{tabular}{lcccccc}
  \toprule
  & \textbf{Gain (dB)} & \textbf{GBW (MHz)} & \textbf{PM (°)} & \textbf{$z_1$ (MHz)} & \textbf{$p_2$ (kHz)} & \textbf{$p_1$ (kHz)} \\
  \midrule
  Theo.& 69.26 & 377.6 & 64.4 & 181.03 & 191.8 & $\sim$120 \\
  Simu. & 69.27 & 8.5 & 6.1 & 181.23 & 459.0 & 54.3 \\
  Error & 0.01\% & 4337\% & 90.5\% & 0.11\% & 58.2\% & $\sim$120\% \\
  \bottomrule
  \end{tabular}%
  }
  \end{center}

  \textbf{Flaw diagnosis}:
  Gain and power were predicted with high accuracy, validating the DC operating point model. 
  Zero $z_1$ prediction was also very precise, confirming the feedforward path modeling.
  However, GBW and PM failed completely. The white-box model predicted a healthy 377 MHz bandwidth and 64.4° PM, while the actual circuit had only 8.5 MHz bandwidth and was nearly unstable (6.1°).
  
  More specifically, the optimizer minimized the compensation capacitor $C_m$ to its lower bound ($10 \text{ fF}$).
  This caused the non-dominant pole $p_2$ to move to a very low frequency ($\approx 190-460 \text{ kHz}$).
  The white-box model calculated $\text{GBW} = G_1 / (2\pi C_m) \approx$377.6 MHz. 
  This formula assumes the system has a single-pole roll-off (-20 dB/dec) up to the unity-gain frequency (GBW).
  However, since $p_2$ (459 kHz) is much lower than the calculated GBW (377 MHz), the gain curve breaks into a -40 dB/dec slope very early. The actual GBW drastically collapsed to 8.5 MHz.
  The PM calculation in the white box used the phase at the calculated GBW, 
  ignoring that the actual crossover occurred much earlier where phase behavior is dominated by the two low-frequency poles rather than one...

  \textbf{Symbolic tracing}:
  The design is unstable (PM = 6.1°) and the predicted GBW is invalid (40x error). 
  The optimizer exploited a model deficiency (lack of explicit pole separation constraint) to produce a mathematically optimal but physically non-functional design. 
  The assumption that $\text{GBW} \approx G_1/(2\pi C_m)$ holds only if $|p_2| > \text{GBW}$. 
  This condition was violated...
  Therefore, we need to reformulate the constraints to strictly enforce the PZ positioning step...

  \textbf{Correction proposal}:
  Introduce a strict extra \textbf{\textit{cons. C3}}: $|p_2| \ge 2 \cdot \text{GBW}$. 
  This will force the optimizer to increase $C_m$ or $G_2$ to push the non-dominant pole out, ensuring the system behaves as a single-pole system up to the crossover frequency...
  \end{examplebox}

\vspace{-.5em}
\begin{examplebox}[Key response 9, iteration 2 —— Dual-criterion checking]
  \begin{center}
  {%
    \setlength{\tabcolsep}{2pt}
    \renewcommand{\arraystretch}{0.7}
    \begin{tabular}{lcccccc}
    \toprule
    & \textbf{Gain (dB)} & \textbf{GBW (MHz)} & \textbf{PM (°)} & \textbf{$z_1$ (MHz)} & \textbf{$p_2$ (kHz)} & \textbf{$p_1$ (kHz)} \\
    \midrule
    Theo. & 67.23 & 1.06 & 65.0 & 38.8 & 2119.7 & $\approx$2.89 \\
    Simu. & 67.24 & 0.94 & 67.0 & 152 & 2180 & 2.79 \\
    Error & 0.01\% & 12.7\% & 3.0\% & 74.5\% & 2.9\% & 3.6\% \\
    \bottomrule
    \end{tabular}%
  }
  \end{center}

ACCEPT... Poles are predicted within 4\% error; the large zero deviation is benign because the zero remains non-dominant (both $\gg$ GBW)...
  \end{examplebox}

\subsubsection{Summary}
In the causality-driven refinement loop,
each refinement iteration rectifies the symbolic design at the reasoning level. 
The output 
includes the optimal design variable $\mathbf{x}^*$, the corresponding theoretical performance metrics $\hat{\mathbf{y}}^*$, the simulation result $\mathbf{y}^*$,
and the entire design trajectory providing complete, interpretable design rationale.

\section{Experiments}\label{s4}

\subsection{Experimental Setup}
\subsubsection{\textbf{Environment}}
\texttt{Gemini-3-pro-preview} \cite{gemini3} serves as the backbone LLM of the agent in White-Op. 
\texttt{HSPICE} simulator performs BL simulation in the loops.
To further validate the reliability of the produced designs, the BL solutions are mapped to TL using the $g_m/I_D$-based tool~\cite{BOtopo3} under a 180-nm process without further sizing.

\subsubsection{\textbf{Baseline and benchmark}}
Black-box methods, GA, BO, and RL, serve as baselines.
Maximum iterations for White-Op are set to 3. 
The specs are: $\text{Gain}>\text{60}\,\text{dB}$, $\text{GBW}>\text{0.5}\,\text{MHz}$, $\text{Power}<\text{250}\,\mu\text{W}$, and $\text{PM}\in[\text{45}^\circ,\text{90}^\circ]$.
Designs with PM outside this range are marked as failures.
Load capacitance $C_L$ is fixed at 10\,pF and load resistance $R_L$ is fixed at 10\,$\text{M}\Omega$. 
Experiments are conducted on 9 op-amp topologies from \cite{review} with 10 trials for each topology in each method.

\subsection{Results and Analysis}
Table~\ref{table:whiteop} shows White-Op's results across the theoretical (Th.) level, BL, and TL, with each row averaging 10 trials.
Tables \ref{table:ga}, \ref{table:bo} and \ref{table:rl} show GA, BO, and RL results respectively. 
On average, all methods take about 15 minutes. 
For GA, time consumption is primarily attributed to simulation; 
for BO, it is mainly attributed to Gaussian process model training; 
for RL, it is mainly attributed to policy model training and simulation; 
for White-Op, it is mainly attributed to LLM reasoning.

White-Op's characteristics are analyzed as follows.

\subsubsection{\textbf{Design effectiveness of White-Op}} 
As shown in Table~\ref{table:whiteop}, the theoretical designs by the agent exhibit an average relative error of only 8.52\% against BL simulation results across all topologies. 
This validates the effectiveness of White-Op in producing designs that closely align with theoretical expectations. 
Note that power is excluded from this comparison, as both the theoretical and BL designs employ the same Eq.~\eqref{eq:power} to estimate power.

\subsubsection{\textbf{Design interpretability of White-Op}} 
As shown in the MZC design example across Section~\ref{s3}, White-Op provides fully interpretable design rationale for human understanding and refinement without black-box sizers or rule-based symbolic derivation tools.
This also explains why Table~\ref{table:whiteop} includes rows for Th. results, whereas Tables \ref{table:ga}, \ref{table:bo} and \ref{table:rl} do not.

Conversely, all black-box baselines, GA, BO and RL lack interpretability. 
Black-box methods yield BL designs with 2.87$\times$ to 11.72$\times$ average FoM compared to White-Op (put aside the reversed situation after TL mapping for the validation of design reliability).
This is expected since White-Op largely prunes the design space via introducing physically meaningful constraints, while GA, BO and RL baselines search the full design space as a black-box. 

\subsubsection{\textbf{Design reliability of White-Op}}
In analog circuit design involving multiple abstract levels, 
the primary goal at the early design phase is not merely to pursue high performance, 
but to implement relatively reliable solutions.
Therefore, to further assess whether the BL designs generated by each method remain valid after downstream implementation, a comparative analysis after TL mapping is conducted. 

Tables \ref{table:whiteop}, \ref{table:ga}, \ref{table:bo}, and \ref{table:rl} show that
the BL$\to$TL performance metric discrepancy for all methods is common and should not be attributed to the proposed White-Op. 
Actually, the performance degradation arises from limitations in BL modeling (Section~\ref{sec:behavioral-level-modeling}),
which captures dominant parasitics $C_{gs}$ but neglects more complex parasitics $C_{gd}$ and $C_{db}$. 
The BL modeling limitations apply equally to all methods under comparison, 
rather than being a shortcoming of White-Op.

On the other hand, White-Op's designs, despite these modeling imperfections, remain valid for all tested topologies after TL mapping. 
This reliability stems from the key idea: 
the clear constraint-introduction reasoning steps in Sections~\ref{sec:approximate-pz-extraction} and \ref{sec:design-from-pzs} inherently enforce design redundancies (e.g., relative PZ positions), ensuring design validity (PM in a reasonable range from 56.12$^\circ$ to 76.52$^\circ$) even under complex parasitic effects. 
For example, in the MZC design example presented in Section~\ref{s3}, when considering $p_2 > \text{GBW}$, White-Op actually introduced $p_2 > 2\cdot\text{GBW}$, thereby proactively incorporating design redundancies at the BL to counteract downstream performance degradation.

\label{sec:experimental_setup}
\begin{table*}[!b]
    \centering
    \caption{Performance comparison among Theory, BL, and TL results using White-Op}
    \label{table:whiteop}
    \setlength{\tabcolsep}{3.5pt}
    \renewcommand{\arraystretch}{1.1}
    \footnotesize
    \begin{tabular}{c|ccc|ccc|ccc|cc|cc|ccc|c}
        \hline
        \multirow{2}{*}{\textbf{Topo.}} & \multicolumn{3}{c|}{\textbf{Gain} (dB)} & \multicolumn{3}{c|}{\textbf{GBW} (MHz)} & \multicolumn{3}{c|}{\textbf{PM} (°)} & \multicolumn{2}{c|}{\textbf{Power} ($\mu$W)} & \multicolumn{2}{c|}{\textbf{Rel. Err.}} & \multicolumn{3}{c|}{\textbf{FoM}} & \textbf{Time} \\
        & Th. & BL & TL & Th. & BL & TL & Th. & BL & TL & Th./BL & TL & Th.$\to$BL & BL$\to$TL & Th. & BL & TL & (min) \\ \hline
        SMC & 63.83 & 63.75 & 79.90 & 3.42 & 3.08 & 1.24 & 58.51 & 62.38 & 65.93 & 16.94 & 48.29 & 6.08\% & 74.27\% & 1135.82 & 1012.37 & 176.54 & 7.66 \\ \hline
        SMCNR & 67.81 & 67.66 & 83.15 & 1.54 & 1.20 & 0.71 & 61.66 & 64.05 & 62.95 & 13.72 & 43.93 & 7.82\% & 80.43\% & 1320.59 & 973.66 & 158.08 & 10.83 \\ \hline
        MZC & 66.33 & 66.35 & 83.41 & 1.22 & 1.04 & 0.61 & 59.79 & 63.73 & 62.12 & 37.39 & 43.71 & 7.09\% & 68.70\% & 794.55 & 679.65 & 138.76 & 10.83 \\ \hline
        NMC & 98.71 & 98.59 & 116.53 & 2.65 & 3.28 & 1.67 & 59.58 & 63.63 & 69.93 & 51.67 & 125.52 & 13.03\% & 67.83\% & 491.26 & 569.96 & 116.30 & 15.39 \\ \hline
        NMCNR & 99.36 & 99.25 & 118.42 & 1.66 & 1.82 & 1.00 & 60.56 & 63.15 & 74.74 & 35.94 & 89.80 & 7.45\% & 61.98\% & 537.70 & 594.63 & 123.93 & 16.90 \\ \hline
        MNMC & 96.76 & 96.00 & 118.88 & 2.79 & 3.30 & 1.80 & 62.76 & 61.15 & 74.04 & 66.55 & 155.86 & 9.22\% & 54.87\% & 462.66 & 556.40 & 124.18 & 16.35 \\ \hline
        NGCC & 103.49 & 103.40 & 116.04 & 1.80 & 1.79 & 1.06 & 62.69 & 58.35 & 71.44 & 41.43 & 103.83 & 5.93\% & 62.36\% & 503.77 & 531.60 & 113.57 & 13.01 \\ \hline
        NMCF & 93.16 & 92.84 & 112.37 & 1.90 & 2.15 & 1.14 & 63.30 & 61.73 & 76.52 & 47.01 & 117.59 & 6.67\% & 75.11\% & 788.54 & 882.88 & 131.49 & 16.05 \\ \hline
        DFCFC & 97.04 & 96.81 & 117.37 & 6.60 & 5.50 & 3.46 & 57.06 & 42.50 & 56.12 & 20.88 & 74.28 & 13.36\% & 93.68\% & 3090.23 & 2547.03 & 439.92 & 17.63 \\ \hline
    \end{tabular}
\end{table*}

\begin{table*}[!b]
    \centering
    \caption{Performance comparison between BL and TL results using GA}
    \label{table:ga}
    \setlength{\tabcolsep}{8.3pt}
    \renewcommand{\arraystretch}{1.1}
    \footnotesize
    \begin{tabular}{c|cc|cc|cc|cc|c|cc|c}
        \hline
        \multirow{2}{*}{\textbf{Topo.}} & \multicolumn{2}{c|}{\textbf{Gain} (dB)} & \multicolumn{2}{c|}{\textbf{GBW} (MHz)} & \multicolumn{2}{c|}{\textbf{PM} (°)} & \multicolumn{2}{c|}{\textbf{Power} ($\mu$W)} & \textbf{Rel. Err.} & \multicolumn{2}{c|}{\textbf{FoM}} & \textbf{Time} \\
        & BL & TL & BL & TL & BL & TL & BL & TL & BL$\to$TL & BL & TL & (min) \\ \hline
        SMC & 70.20 & 77.08 & 3.35 & 1.85 & 55.01 & 61.17 & 45.42 & 106.79 & 54.47\% & 785.05 & 175.10 & 14.61 \\ \hline
        SMCNR & 70.73 & 71.68 & 7.84 & 4.31 & 55.35 & 79.94 & 5.58 & 30.33 & 134.41\% & 14076.14 & 1420.27 & 14.76 \\ \hline
        MZC & 60.09 & 64.41 & 14.48 & 8.67 & 55.15 & 33.05 & 93.78 & 188.26 & 47.91\% & 1534.66 & fail & 13.62 \\ \hline
        NMC & 103.02 & 116.15 & 6.02 & 4.02 & 64.67 & -116.92 & 3.49 & 28.57 & 262.18\% & 17235.16 & fail & 15.82 \\ \hline
        NMCNR & 98.45 & 116.33 & 25.03 & 14.15 & 60.68 & 131.74 & 9.16 & 43.17 & 144.80\% & 30091.91 & fail & 15.26 \\ \hline
        MNMC & 98.43 & 110.13 & 8.68 & 6.68 & 63.77 & -116.57 & 6.70 & 36.48 & 193.05\% & 13056.89 & fail & 14.83 \\ \hline
        NGCC & 98.16 & 103.70 & 18.01 & 15.83 & 64.37 & -105.62 & 22.30 & 74.75 & 135.96\% & 8626.38 & fail & 13.97 \\ \hline
        NMCF & 97.94 & 107.01 & 12.44 & 10.13 & 64.44 & -109.17 & 12.05 & 51.10 & 158.84\% & 10479.67 & fail & 13.45 \\ \hline
        DFCFC & 102.25 & 99.62 & 28.26 & 7.14 & 59.12 & 77.06 & 143.33 & 367.66 & 65.60\% & 1975.65 & 199.75 & 14.80 \\ \hline
    \end{tabular}
\end{table*}

\begin{table*}[!b]
    \centering
    \caption{Performance comparison between BL and TL results using BO}
    \label{table:bo}
    \setlength{\tabcolsep}{8.3pt}
    \renewcommand{\arraystretch}{1.1}
    \footnotesize
    \begin{tabular}{c|cc|cc|cc|cc|c|cc|c}
        \hline
        \multirow{2}{*}{\textbf{Topo.}} & \multicolumn{2}{c|}{\textbf{Gain} (dB)} & \multicolumn{2}{c|}{\textbf{GBW} (MHz)} & \multicolumn{2}{c|}{\textbf{PM} (°)} & \multicolumn{2}{c|}{\textbf{Power} ($\mu$W)} & \textbf{Rel. Err.} & \multicolumn{2}{c|}{\textbf{FoM}} & \textbf{Time} \\
        & BL & TL & BL & TL & BL & TL & BL & TL & BL$\to$TL & BL & TL & (min) \\ \hline
        SMC & 70.20 & 76.28 & 3.58 & 2.02 & 57.66 & 57.91 & 50.00 & 117.03 & 62.17\% & 969.01 & 190.64 & 12.02 \\ \hline
        SMCNR & 68.80 & 64.72 & 15.59 & 6.86 & 57.82 & 85.02 & 20.13 & 53.84 & 87.99\% & 9470.06 & 1083.13 & 11.82 \\ \hline
        MZC & 63.56 & 58.91 & 13.42 & 6.81 & 58.25 & 53.44 & 145.78 & 279.60 & 43.34\% & 952.28 & 238.31 & 11.89 \\ \hline
        NMC & 103.36 & 97.50 & 34.05 & 29.71 & 59.79 & -108.35 & 123.53 & 320.46 & 116.73\% & 2804.09 & fail & 13.57 \\ \hline
        NMCNR & 105.07 & 98.24 & 49.49 & 31.25 & 59.62 & -167.76 & 143.07 & 347.34 & 141.95\% & 3548.60 & fail & 13.34 \\ \hline
        MNMC & 101.62 & 95.65 & 27.51 & 21.78 & 59.74 & -113.28 & 129.35 & 348.28 & 125.06\% & 2107.47 & fail & 13.21 \\ \hline
        NGCC & 99.88 & 95.34 & 27.76 & 23.60 & 58.78 & -111.64 & 182.47 & 481.02 & 120.29\% & 1536.34 & fail & 13.37 \\ \hline
        NMCF & 101.81 & 96.64 & 32.48 & 25.73 & 59.76 & -105.75 & 171.93 & 464.98 & 118.31\% & 1910.40 & fail & 13.58 \\ \hline
        DFCFC & 100.36 & 97.55 & 12.17 & 5.99 & 57.92 & 67.97 & 180.45 & 465.37 & 57.70\% & 678.70 & 127.53 & 14.88 \\ \hline
    \end{tabular}
\end{table*}
\begin{table*}[!b]
    \centering
    \caption{Performance comparison between BL and TL results using RL}
    \label{table:rl}
    \setlength{\tabcolsep}{8.3pt}
    \renewcommand{\arraystretch}{1.1}
    \footnotesize
    \begin{tabular}{c|cc|cc|cc|cc|c|cc|c}
        \hline
        \multirow{2}{*}{\textbf{Topo.}} & \multicolumn{2}{c|}{\textbf{Gain} (dB)} & \multicolumn{2}{c|}{\textbf{GBW} (MHz)} & \multicolumn{2}{c|}{\textbf{PM} (°)} & \multicolumn{2}{c|}{\textbf{Power} ($\mu$W)} & \textbf{Rel. Err.} & \multicolumn{2}{c|}{\textbf{FoM}} & \textbf{Time} \\
        & BL & TL & BL & TL & BL & TL & BL & TL & BL$\to$TL & BL & TL & (min) \\ \hline
        SMC & 71.72 & 64.00 & 20.54 & 11.90 & 59.01 & 18.37 & 115.10 & 229.66 & 56.88\% & 1999.16 & fail & 15.29 \\ \hline
        SMCNR & 68.97 & 71.04 & 8.05 & 4.30 & 60.26 & 83.08 & 6.59 & 31.88 & 123.70\% & 12181.25 & 1331.98 & 15.51 \\ \hline
        MZC & 66.52 & 62.73 & 19.34 & 11.25 & 58.56 & 22.95 & 121.72 & 236.48 & 51.68\% & 1601.13 & fail & 15.01 \\ \hline
        NMC & 106.74 & 99.69 & 31.57 & 25.98 & 58.78 & -101.43 & 84.09 & 236.12 & 120.03\% & 3795.85 & fail & 15.16 \\ \hline
        NMCNR & 104.48 & 100.38 & 45.71 & 29.89 & 61.36 & -137.24 & 83.90 & 220.06 & 141.59\% & 5652.37 & fail & 14.70 \\ \hline
        MNMC & 101.88 & 94.41 & 32.05 & 27.81 & 61.67 & -105.37 & 112.02 & 278.65 & 112.19\% & 2931.95 & fail & 15.08 \\ \hline
        NGCC & 99.20 & 93.50 & 31.51 & 25.51 & 59.15 & -101.22 & 170.87 & 436.18 & 113.59\% & 1888.04 & fail & 15.66 \\ \hline
        NMCF & 103.49 & 97.12 & 30.11 & 22.39 & 59.59 & -101.15 & 126.62 & 319.96 & 114.96\% & 2407.13 & fail & 15.03 \\ \hline
        DFCFC & 100.84 & 97.67 & 30.19 & 10.60 & 58.79 & 67.11 & 174.66 & 458.62 & 61.50\% & 1686.30 & 236.13 & 15.69 \\ \hline
    \end{tabular}
\end{table*}

All black-box baselines' seemingly better BL results, 
however, suffer severe degradation after TL mapping. 
For GA, BO, and RL, in 5 to 7 out of 9 op-amp topologies, PM consistently falls outside reasonable ranges (averaging from $\text{-39.43}^\circ$ to $\text{-7.26}^\circ$), rendering these circuits invalid.

This contrasting outcome stems from their mechanisms: black-box methods use a larger design space and a naive high-FoM objective, which leads to BL designs with highly entangled, unsafe PZ locations that are fragile in downstream implementation.
White-Op's interpretability maintains redundancy, ensuring reliability.

Besides, black-box baselines do yield valid TL results for a few topologies, but such narrow success can largely be attributed to coincidental parasitic tolerance of PZ positions in these specific topologies. 
Even among these cases, warning signs persist: 
e.g., 
For GA, the MZC op-amp exhibits a PM of $\text{33.05}^\circ$ at TL, posing a severe stability risk as it easily fails into oscillation across temperature and manufacturing variations. 
For RL, not only does MZC suffer from inadequate PM, but even the basic SMC circuit shows a drastically low PM of just $\text{18.37}^\circ$ at TL, exhibiting a non-functional, under-damped unstable state.
Conversely, for BO, the SMCNR op-amp gives a PM of $\text{85.02}^\circ$ at TL, exhibiting an over-damped state that is equally undesirable due to the severe sacrifice in transient response speed.
These examples indicate that BL designs generated by black-box methods are not only uninterpretable, but also unreliable after TL mapping.

\section{Conclusions}\label{s5}
% overview
This paper presents an op-amp BL design framework based on agentic human-mimicking reasoning.
A reasoning-solving decoupled paradigm guides the agent to perform symbolic reasoning and formulate an optimization problem that is solved programmatically, verified via simulation, and iteratively refined. 
Introducing hypothetical constraints during 
TF simplification, PZ extraction, and positioning transforms implicit reasoning into explicit tasks.
A programming mapping protocol enables stable translation from symbolic designs to executable programs.
A causality-driven refinement loop enables the agent to trace design flaws back to specific steps and correct them.
White-Op achieves interpretable BL designs while retaining functionality after TL mapping.
%

%\section*{Acknowledgments}
%This paper uses the generative AI (Claude series)
%only for grammar checking and readability improvement.

\bibliographystyle{IEEEtran}
\bibliography{reference.bib}

\newpage
\begin{IEEEbiography}[{\includegraphics[width=1in,height=1.25in,clip,keepaspectratio]{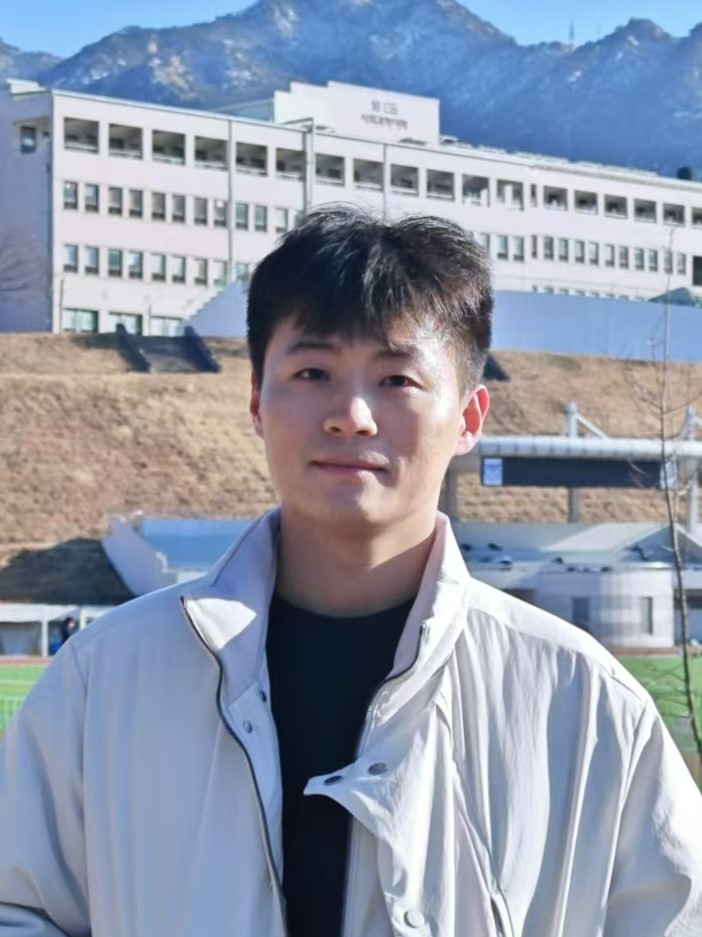}}]
{Zihao Chen} received the B.E. degree from the School of Microelectronics at Fudan University, Shanghai, China, in 2021 and is currently pursuing the Ph.D. at the same institution. His research focuses on the cutting-edge technologies like reinforcement learning (RL) and large language models (LLMs), with their applications in electronic design automation (EDA).
\end{IEEEbiography}

\begin{IEEEbiography}[{\includegraphics[width=0.95in,height=1.2in,clip,keepaspectratio]{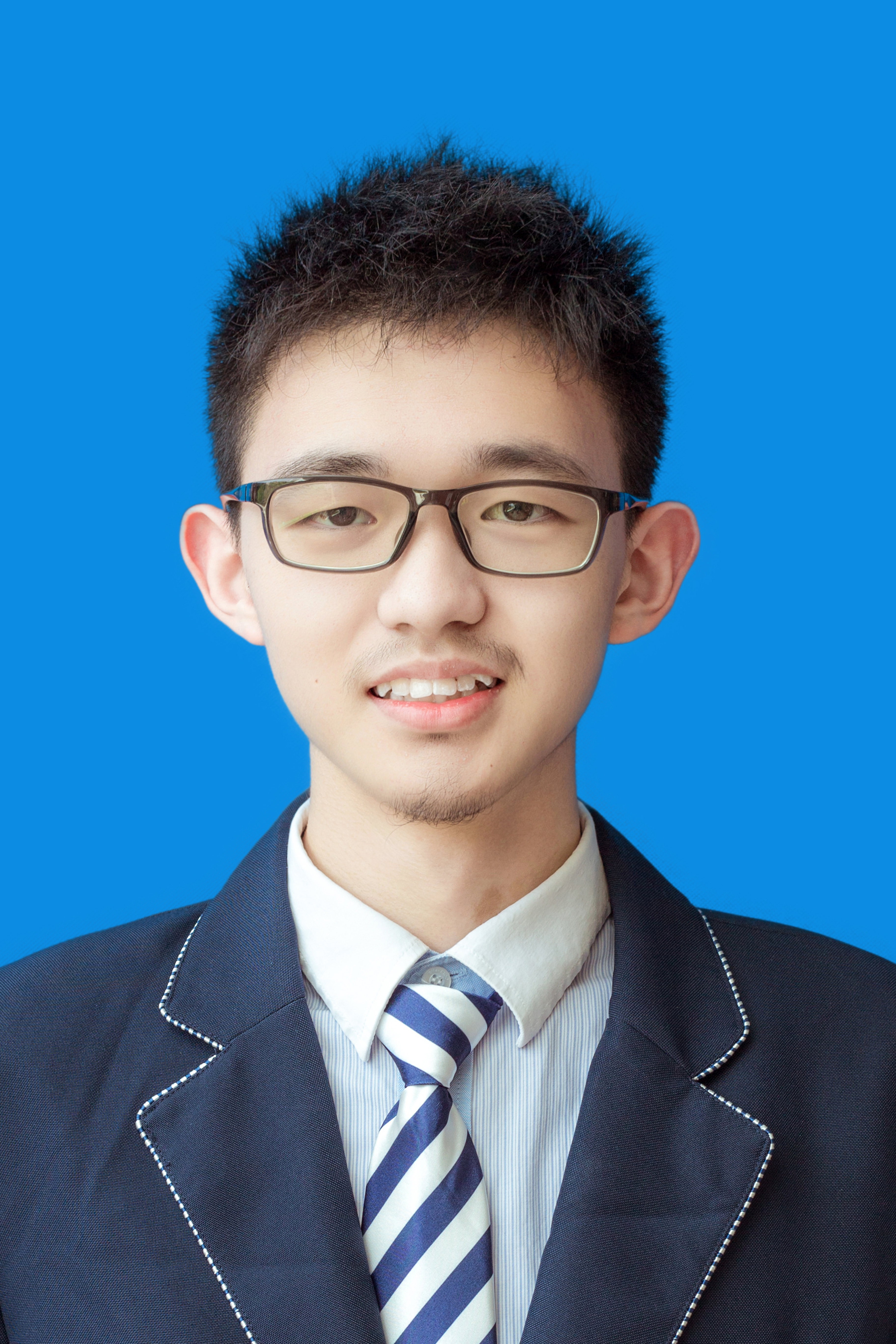}}]
{Ziyi Sun} is currently pursuing a bachelor's degree in Microelectronics Science and Engineering at Fudan University, Shanghai, China. His current research focuses on the application of large language models in EDA.
\end{IEEEbiography}

\begin{IEEEbiography}[{\includegraphics[width=0.95in,height=1.2in,clip,keepaspectratio]{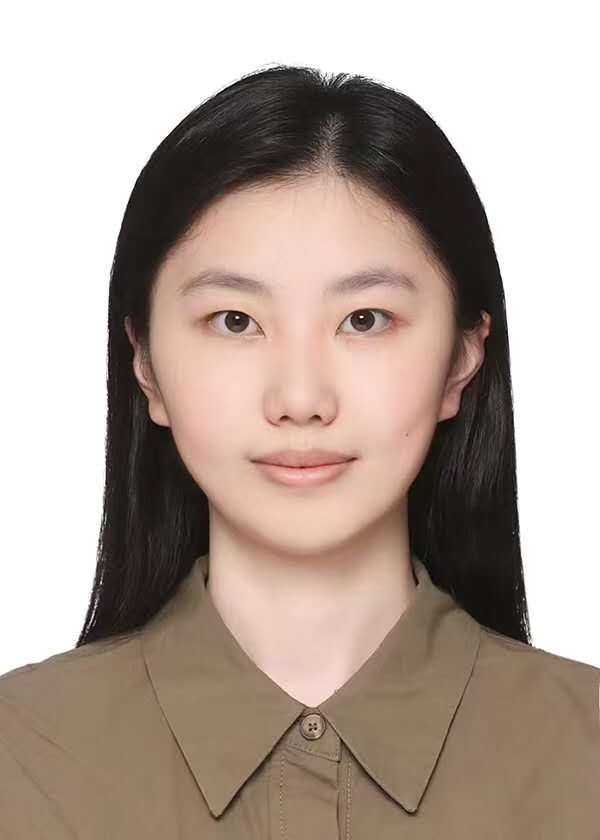}}]
{Jiayin Wang} is currently pursuing a bachelor's degree in Microelectronics Science and Engineering at Fudan University, Shanghai, China. Her current research focuses on analog circuits and large language models.
\end{IEEEbiography}

\begin{IEEEbiography}[{\includegraphics[width=0.95in,height=1.2in,clip,keepaspectratio]{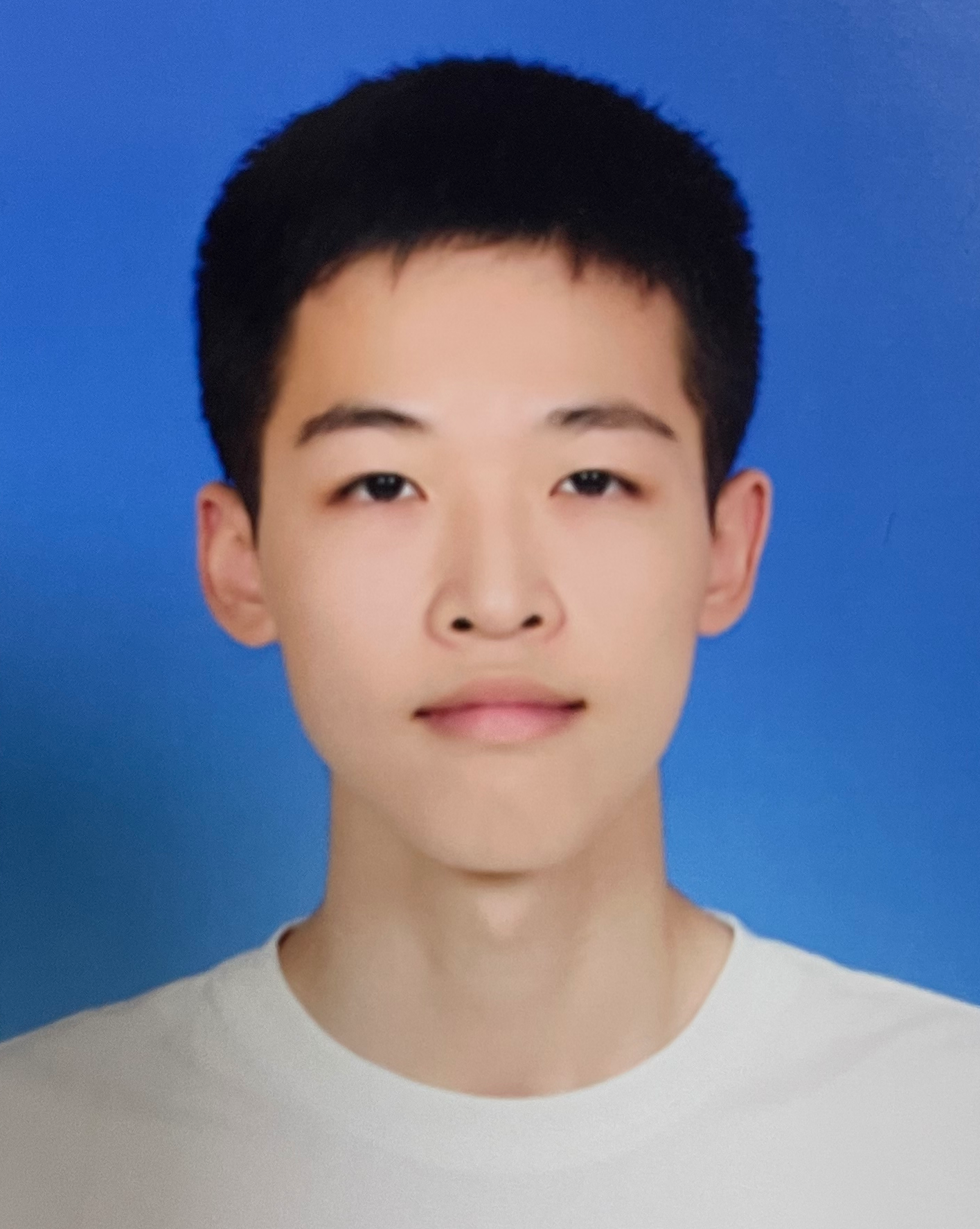}}]
{Ji Zhuang} received the B.E. degree from the School of Microelectronics at Fudan University, Shanghai, China, in 2025 and is currently pursuing the Ph.D. at the same institution. His research interests lie in EDA, with a focus on leveraging LLMs to advance automation in digital front-end design and analog circuit design.
\end{IEEEbiography}

\begin{IEEEbiography}[{\includegraphics[width=1in,height=1.25in,clip,keepaspectratio]{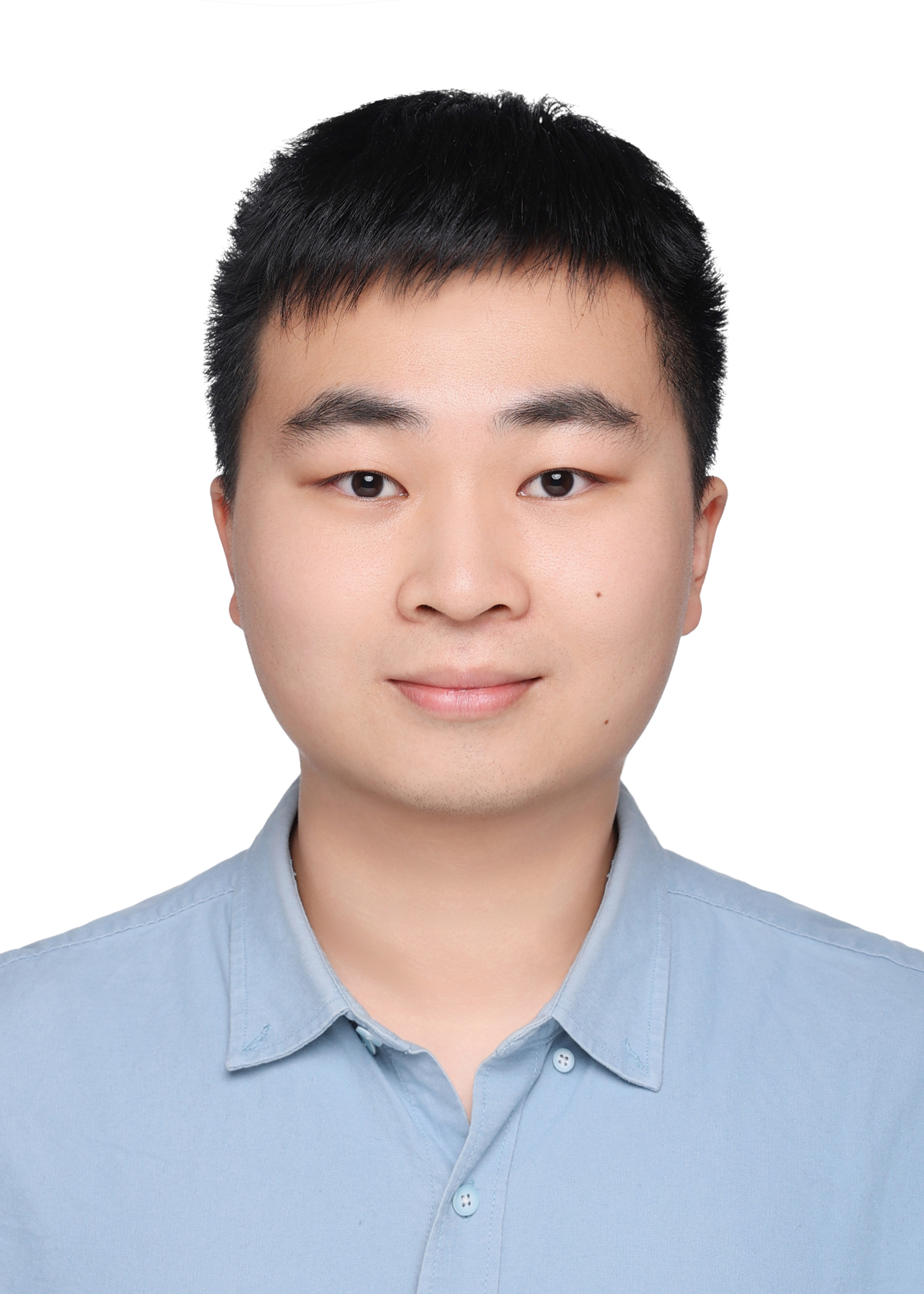}}]
  {Jinyi Shen} 
   received the B.S. degree in microelectronics from Fudan University, Shanghai, China, in 2021, where he is currently pursuing the Ph.D. degree with the School of Microelectronics. His current research interests include analog circuit design automation and AI for EDA.
\end{IEEEbiography}

\begin{IEEEbiography}[{\includegraphics[width=1in,height=1.25in,clip,keepaspectratio]{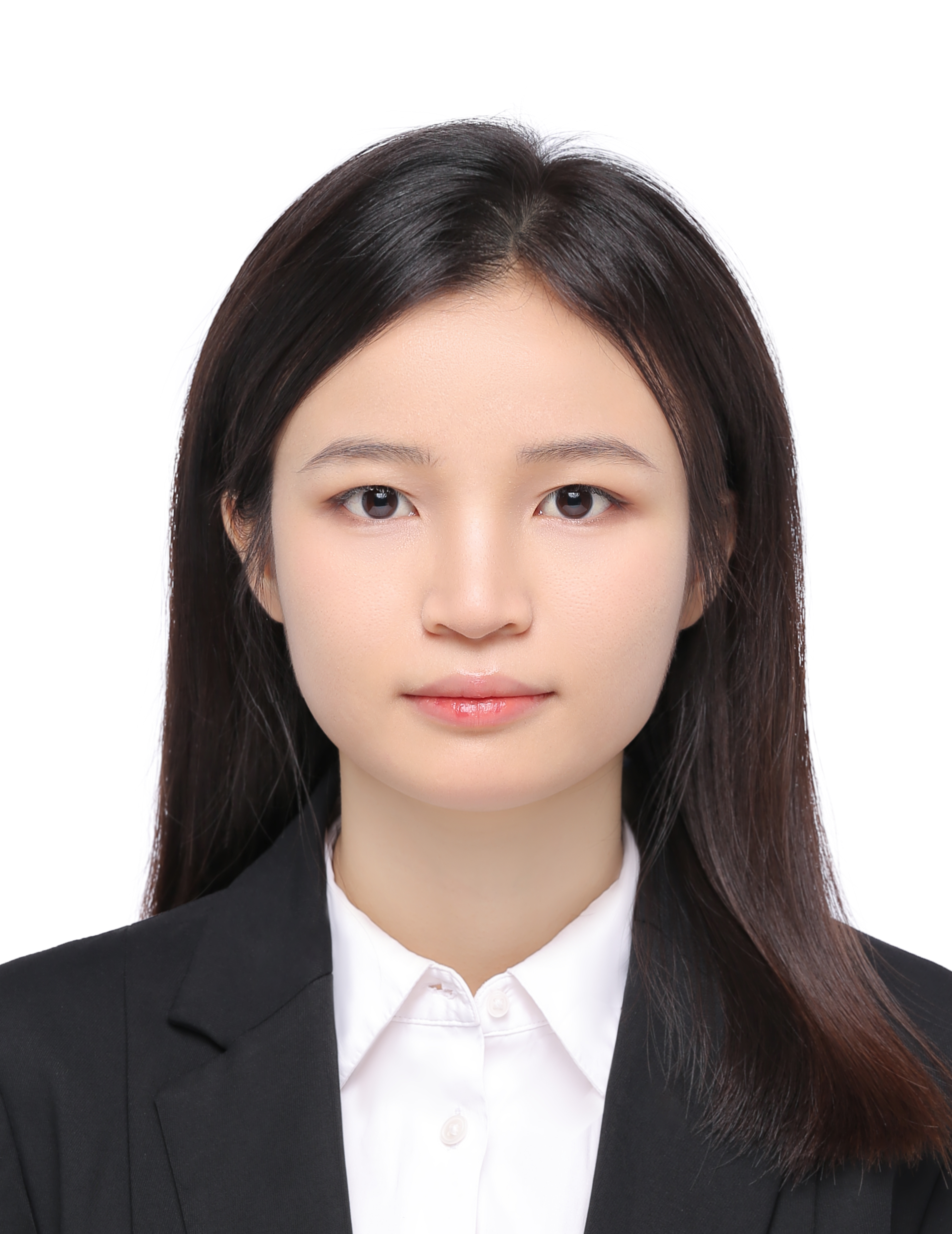}}]
    {Xiaoyue Ke} received the B.S. degree in microelectronics from Fudan University, Shanghai, China, in 2024. She is currently working toward the M.S. degree in microelectronics at the same university. Her research interests include analog integrated circuit design, particularly in the area of high-speed SerDes interfaces.
\end{IEEEbiography}

\begin{IEEEbiography}
	[{\includegraphics[width=1in,height=1.25in,clip,keepaspectratio]{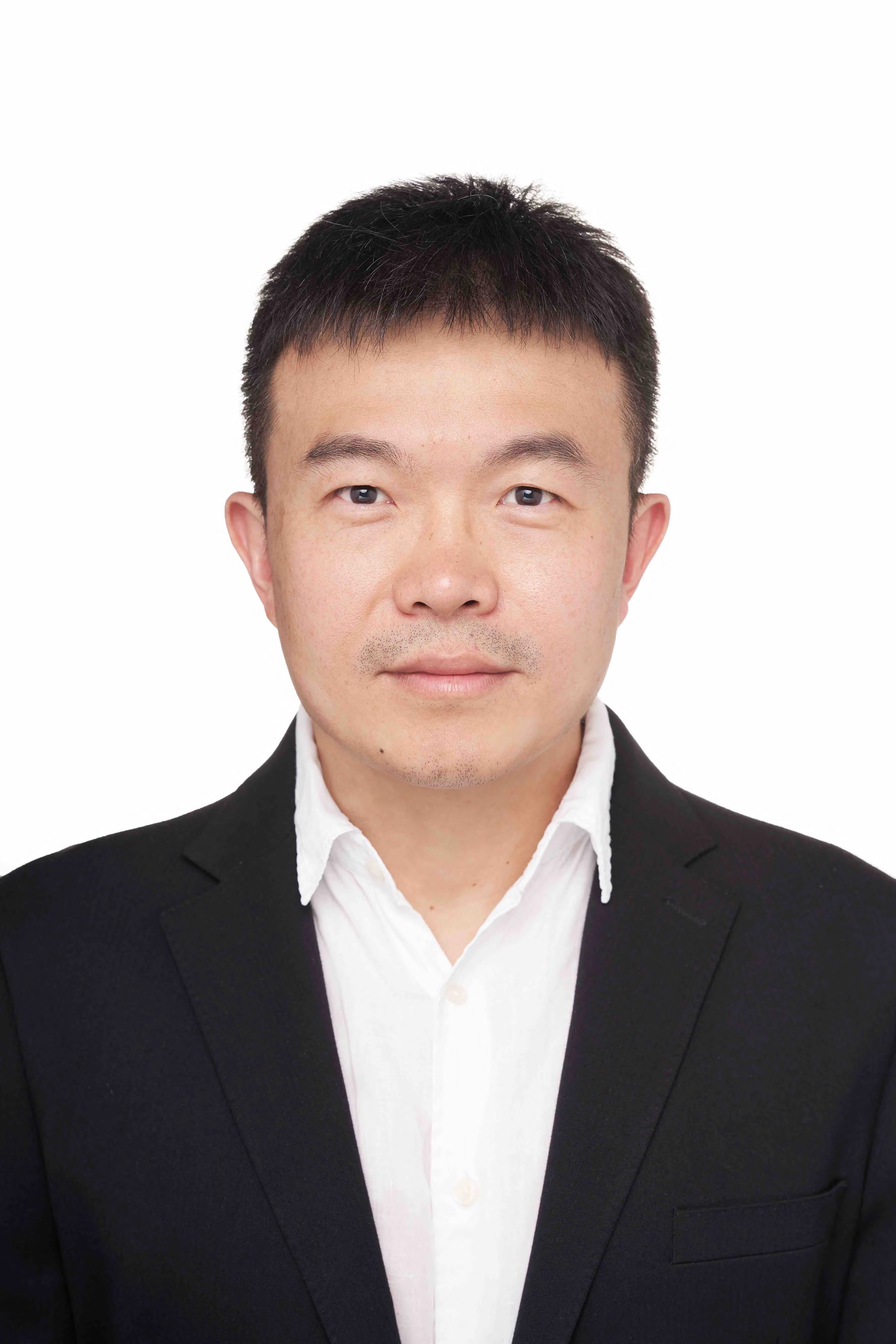}}]
	{Li Shang} is a Professor at the School of Computer Science, Fudan University. He received his Ph.D. degree from Princeton University. He was the Deputy Director and Chief Architect of Intel Labs China, and Associate Professor at the University of Colorado Boulder. His research focuses on embodied intelligence, machine learning, and VLSI \& EDA. He has over 170 publications with multiple best paper awards and nominations, and over 8000 citations. He was a recipient of the NSF Career Award.
\end{IEEEbiography}

\begin{IEEEbiography}[{\includegraphics[width=1in,height=1.25in,clip,keepaspectratio]{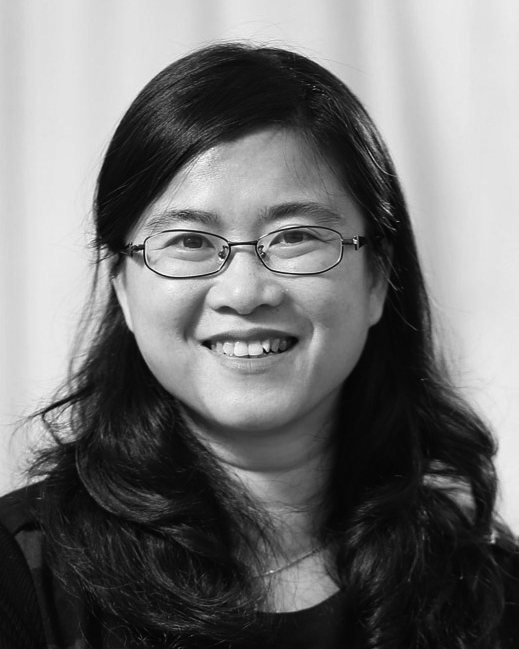}}]
  {Xuan Zeng} received the B.S. and Ph.D. degrees in electrical engineering from Fudan University, Shanghai, China, in 1991 and 1997, respectively. 
  She is a Full Professor with School of Microelectronics, Fudan University. 
  Her research focuses on analog circuit modeling and synthesis, design for manufacturability, high-speed interconnect analysis and optimization, and circuit simulation. 
  She received the Changjiang Distinguished Professor with the Ministry of Education Department of China in 2014. 
\end{IEEEbiography}

\begin{IEEEbiography}
	[{\includegraphics[width=1in,height=1.25in,clip,keepaspectratio]{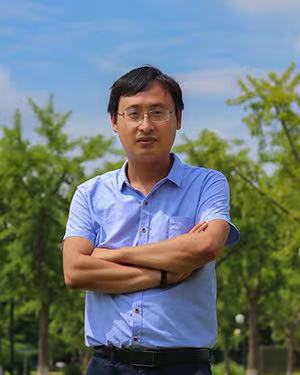}}]
	{Fan Yang} received B.S. degree from Xi'an Jiaotong University in 2003 and Ph.D. degree from Fudan University, Shanghai, China, in 2008. He is a Full Professor with the Microelectronics Department in Fudan University.
	His research interests include model order
	reduction, circuit simulation, high-level synthesis, acceleration of artificial neural networks,
	yield analysis and design for manufacturability.
\end{IEEEbiography}

\end{document}